\documentclass{article} 
\usepackage{iclr2021_conference,times}

\usepackage{amsmath,amsfonts,bm}

\def\eqref#1{equation~\ref{#1}}

\def\1{\bm{1}}

\def\vp{{\bm{p}}}

\def\vu{{\bm{u}}}

\def\vw{{\bm{w}}}

\def\vz{{\bm{z}}}

\def\mN{{\bm{N}}}

\DeclareMathAlphabet{\mathsfit}{\encodingdefault}{\sfdefault}{m}{sl}
\SetMathAlphabet{\mathsfit}{bold}{\encodingdefault}{\sfdefault}{bx}{n}

\newcommand{\R}{\mathbb{R}}

\newcommand{\KL}{D_{\mathrm{KL}}}

\usepackage{hyperref}
\usepackage{url}

\usepackage{booktabs}
\usepackage{graphicx}
\usepackage{bm}
\usepackage{subcaption}
\usepackage{color,xcolor}
\usepackage{pifont}
\usepackage{caption}
\usepackage{wrapfig}
\usepackage{algorithm, algpseudocode}
\usepackage{anyfontsize}

\title{Measuring Uncertainty through Bayesian Learning of Deep Neural Network Structure}

\iclrfinalcopy

\author{Zhijie Deng, Yucen Luo \& Jun Zhu \\
{\fontsize{9}{11}\selectfont Dept. of Comp. Sci. and Tech., BNRist Center, Institute for AI, Tsinghua-Bosch Joint ML Center, THBI Lab}\\
{\fontsize{9}{11}\selectfont Tsinghua University, Beijing, 100084, China}\\
{\small \texttt{\{dzj17, luoyc15\}@mails.tsinghua.edu.cn, dcszj@tsinghua.edu.cn}}
}

\let\oldReturn\Return
\renewcommand{\Return}{\State\oldReturn}

\newcommand{\red}[1]{\textcolor{black}{#1}}
\newcommand{\cmark}{\ding{51}}%
\newcommand{\xmark}{\ding{55}}%

\newcommand{\RN}[1]{%
	\textup{\lowercase\expandafter{\it \romannumeral#1}}%
}
\newcommand*{\Scale}[2][4]{\scalebox{#1}{$#2$}}%

\begin{document}

\maketitle
\vspace{-2ex}
\begin{abstract}
\vspace{-2ex}
Bayesian neural networks (BNNs) augment deep networks with uncertainty quantification by Bayesian treatment of the network weights. However, such models face the challenge of Bayesian inference in a high-dimensional and usually over-parameterized space. 
This paper investigates a new line of Bayesian deep learning by performing Bayesian inference on network structure. Instead of building structure from scratch inefficiently, we draw inspirations from neural architecture search to represent the network structure. 
We then develop an efficient stochastic variational inference approach which unifies the learning of both network structure and weights. 
Empirically, our method exhibits competitive predictive performance while preserving the benefits of Bayesian principles across challenging scenarios.
We also provide convincing experimental justification for our modeling choice. 
\end{abstract}

\vspace{-0.3cm}
\section{Introduction}
\vspace{-0.1cm}
One core goal of Bayesian deep learning is to equip the expressive deep neural networks~(DNNs) with appropriate uncertainty quantification.
In this spirit, Bayesian neural networks (BNNs) perform Bayesian inference over network weights to model uncertainty~\citep{graves2011practical,blundell2015weight,gal2016dropout}.
Despite wide adoption, BNNs are not problemless:
on one side, it is hard to specify an effective prior or to employ a flexible variational for the high-dimensional weights~\citep{sun2018functional,pearce2019expressive}, and hence BNNs usually exhibit unsatisfactory performance;
on the other side, diverse weight samples from the weight posterior are likely to predict similarly~\citep{fort2019deep}, resulting in meaningless weight uncertainty, due to the over parameterization of DNNs.

Given a neural network model, apart from weights, the network structure will also affect the predictive behaviour of the model, even more prominently. 
Besides, as shown in neural architecture search (NAS)~\citep{zoph2016neural,pham2018efficient,liu2018darts,cai2018proxylessnas}, the network structure can be defined in a compact manner, \emph{e.g.}, a set of scalars harnessing the information flow among hidden feature maps. 
Therefore, opportunities arise where we can learn the posterior over the compact, low-dimensional network structure for characterizing uncertainty.
Moreover, with Bayesian structure learning equipped, the model perhaps gains extra benefits including boosted predictive performance due to adapting structure \emph{w.r.t.} data, and diversified posterior ensemble thanks to the global and high-level nature of structure.

To these ends, we attempt to develop a Bayesian structure learning paradigm for deep neural networks, named \emph{Deep Bayesian Structure Networks} (DBSN), to conjoin the benefits from Bayesian principles and structure learning. 
Leveraging the compact, low-dimensional network structure representation developed in modern NAS~\citep{liu2018darts,xie2018snas}, 
we build a unified Bayesian learning framework for both the structure and the weights in light of stochastic variational inference.
To bypass the aforementioned obstacles of explicitly maintaining a distribution over weights, we approximately perform maximum a posteriori (MAP) estimation on network weights, which is theoretically sound and differentiates us from classic BNNs.
Empirically, we perform extensive and fair comparisons to validate the effectiveness of DBSN.

\vspace{-0.cm}
\section{Deep Bayesian Structure Networks (DBSN)} 
\vspace{-0.cm}
We at first briefly discuss how to represent the structures of deep neural networks to enable compact structure learning, and then
elaborate DBSN.


\begin{figure}[!t]
\vspace{-0.2cm}
\centering
\begin{subfigure}{0.9\textwidth}
  \centering
  \includegraphics[width=\linewidth]{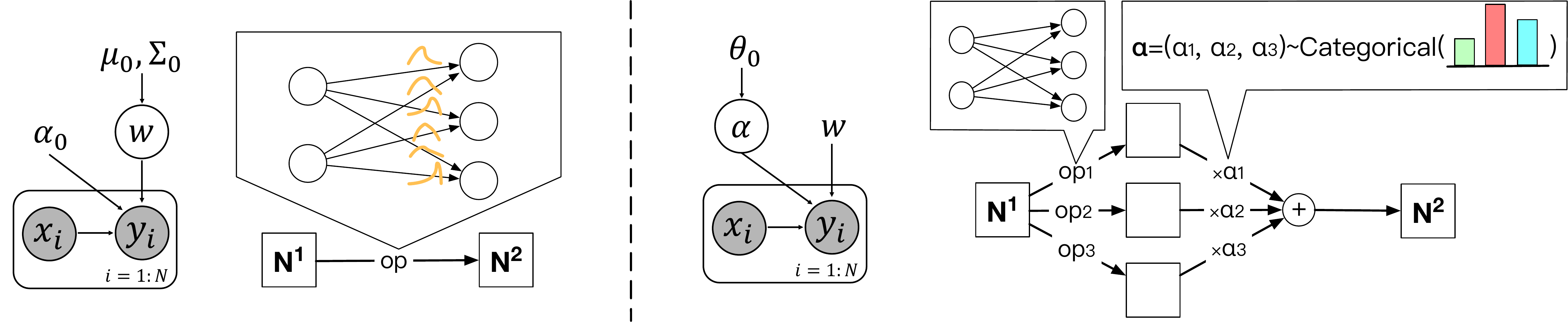}
  \label{fig:illu-1}
\end{subfigure}
\vspace{-0.6cm}
\caption{\red{\footnotesize BNNs with uncertainty on the network weights (left) vs. DBSN with uncertainty on the network structure (right). $\vw$ and $\boldsymbol{\alpha}$ represent network weights and structure, respectively. In DBSN, $\vw$ is also learnable.}}\vspace{-0.3cm}
\label{fig:illu}
\vspace{-0.1cm}
\end{figure}



\subsection{The Structure Representation in DBSN} 
Instead of building network structures from scratch inefficiently~\citep{adams2010learning}, we advocate inheriting the well-evaluated designs in modern NAS~\citep{liu2018darts} to concisely represent the structure. 
Concretely, we construct a network by stacking a sequence of computation cells which have the same internal structure and are separated by downsampling modules. 
Every cell contains $B$ sequential nodes (\emph{i.e.}, feature maps): $\mN^1,\dots,\mN^B$. 
Each node $\mN^j$ is connected to all of its predecessors $\mN^i$ (\emph{i.e.}, $i<j$) by $K$ redundant operations $o^{(i,j)}_1,\dots,o^{(i,j)}_K$.
We define the basic component of the structure as the discrete mask on the $K$ available operations from $\mN^i$ to $\mN^j$, \emph{i.e.}, $\boldsymbol{\alpha}^{(i,j)} \in \{0,1\}^K, \sum_k\boldsymbol{\alpha}^{(i,j)}_k=1$.
The whole structure is defined as $\boldsymbol{\alpha} = \{\boldsymbol{\alpha}^{(i,j)}| 1\leq i<j\leq B\}$. 
In this sense, we can formulate the information flow inside the cell as ($\vw$ as the network weights): 
\vspace{-0.2cm}
\begin{equation}
\label{eq:arch}
\small
    \mN^{(i,j)} = \sum_{k=1}^K \boldsymbol{\alpha}^{(i,j)}_k \cdot o^{(i,j)}_k(\mN^i; \vw), \;\;\;\;\; \mN^j = \sum_{i<j} \mN^{(i,j)}.
\end{equation}

\vspace{-0.5cm}
\subsection{The Theoretical Framework of DBSN}
\vspace{-0.cm}
Formally, let $\mathcal{D}=\{(x_i, y_i)
\}_{i=1}^N$ be a set of $N$ data points, where $x_i \in \mathbb{R}^d$ and $y_i \in\mathcal{Y}$. 
Denoting the data likelihood dictated by the network with weights $\vw$ and structure $\boldsymbol{\alpha}$ as $p(\mathcal{D}|\boldsymbol{\alpha}, \vw)$, we assume a factorized parameter prior $p(\boldsymbol{\alpha}, \vw)=p(\boldsymbol{\alpha})p(\vw)$ and aim at inferring the  posterior $p(\boldsymbol{\alpha}, \vw|\mathcal{D})$.
Directly deriving the posterior is intractable because it is impossible to analytically integrate \emph{w.r.t.} $\boldsymbol{\alpha}$ and $\vw$ for non-linear deep models. 
As suggested by variational BNNs~\citep{graves2011practical,blundell2015weight}, we can alternatively deploy a variational $q(\boldsymbol{\alpha}, \vw)\in \mathcal{Q}$ to approximate $p(\boldsymbol{\alpha}, \vw|\mathcal{D})$.
The posterior inference is accomplished by minimizing the below Kullback-Leibler (KL) divergence:
\begin{equation}
\footnotesize
    \label{eq:elbo}
    \begin{split}
    \Scale[0.95]{
        \min_{q\in \mathcal{Q}} \KL(q(\boldsymbol{\alpha}, \vw)\Vert p(\boldsymbol{\alpha}, \vw|\mathcal{D})) = - \mathbb{E}_{q(\boldsymbol{\alpha}, \vw)}[\log p(\mathcal{D}|\boldsymbol{\alpha}, \vw)] + \KL(q(\boldsymbol{\alpha}, \vw)\Vert p(\boldsymbol{\alpha}, \vw)) + constant.}
    \end{split}
\end{equation}
To avoid the barrier of learning distribution over $\vw$, we untie the entanglement between $\boldsymbol{\alpha}$ and $\vw$ in the variational, and let $\vw$ follow a delta distribution, 
namely, $q(\boldsymbol{\alpha}, \vw)=q(\boldsymbol{\alpha}|\boldsymbol{\theta})\delta(\vw-\vw_0)$ with $\{\boldsymbol{\theta}, \vw_0\}$ as learnable parameters. 
Intuitively, we perform maximum a posteriori (MAP) estimation on $\vw$ to realise empirical Bayesian learning.
Under a common assumption that the prior of $\vw$ is a standard Gaussian, the KL divergence between the variational and prior of $\vw$ reduces to the L2 regularization $\min_{\vw_0}\gamma||\vw_0||^2_2$ with hyper-parameter $\gamma$.
We then write Eq.~(\ref{eq:elbo}) into a practical loss:
\begin{equation}
    \label{eq:loss_alpha}
    \small
    \begin{split}
        \mathcal{L}(\boldsymbol{\theta}, \vw_0) = - \mathbb{E}_{q(\boldsymbol{\alpha}|\boldsymbol{\theta})}[\log p(\mathcal{D}|\boldsymbol{\alpha}, \vw_0)] + \KL(q(\boldsymbol{\alpha}|\boldsymbol{\theta})\Vert p(\boldsymbol{\alpha})) + \gamma||\vw_0||^2_2.
    \end{split}
\end{equation}
Given the composable and discrete nature of the structure, we factorize $q(\boldsymbol{\alpha}|\boldsymbol{\theta})$ to be a multiplication of categorical distributions, \emph{i.e.}, $q(\boldsymbol{\alpha}|\boldsymbol{\theta})=\prod_{i<j}q(\boldsymbol{\alpha}^{(i,j)}|\boldsymbol{\theta}^{(i,j)})$, where $\boldsymbol{\theta} = \{\boldsymbol{\theta}^{(i,j)}\in {\R}^{K}| 1\leq i<j\leq B\}$ 
denote the categorical logits.
Without losing generality, we assume an independent prior: $p(\boldsymbol{\alpha})=\prod_{i<j}p(\boldsymbol{\alpha}^{(i,j)})$ with $p(\boldsymbol{\alpha}^{(i,j)})$ as categorical distributions with uniform class probabilities.
The graphical model is depicted in Figure~\ref{fig:illu}. We denote $\vw_0$ as $\vw$ thereafter if there is no misleading.

To minimize $\mathcal{L}$ \emph{w.r.t.} $\boldsymbol{\theta}$ by gradient, a typical recipe is to relax $q(\boldsymbol{\alpha}^{(i,j)}|\boldsymbol{\theta}^{(i,j)})$ and $p(\boldsymbol{\alpha}^{(i,j)})$ to be concrete distributions to enable the use of reparameterization~\citep{maddison2016concrete}. 
For a stochastic estimation of $ \mathcal{L}$, we can draw samples from $q(\boldsymbol{\alpha}|\boldsymbol{\theta})$ via
{\small $
    \boldsymbol{\alpha}=g(\boldsymbol{\theta},\boldsymbol{\epsilon}) = \{\mathrm{softmax}({(\boldsymbol{\theta}^{(i,j)} + \boldsymbol{\epsilon}^{(i,j)})}/{\tau})\},
$}
where {\small $\boldsymbol{\epsilon} = \{\boldsymbol{\epsilon}^{(i,j)}\in \R^K\}$} are \emph{i.i.d.} Gumbel variables, with $\tau \in \R_+$ as temperature.

Recall that DBSN factorizes the weights and the structure in the variational, so, intuitively, the weights $\vw$ are structure-agnostic and shared among diverse structure samples. 
Considering the limited capacity of the shared weights, we reasonably conjecture that the structure stochasticity induced by the above sampling procedure may pose too strong regularization effects on the weights, giving rise to unsatisfactory weight convergence.
As a remedy, in analogy to the ``cold posterior'' trick~\citep{wenzel2020good}, we propose to sharpen the approximate posterior $q(\boldsymbol{\alpha}|\boldsymbol{\theta})$. 
To this end, we develop the \emph{sharpened concrete distribution} by augmenting the sampling procedure of concrete distribution with a tunable factor $\boldsymbol{\beta}^{(i,j)}$:
\begin{equation}
\vspace{-0.1cm}
    \label{eq:concrete}
    \small
    \begin{split}
    \boldsymbol{\alpha}^{(i,j)}=g(\boldsymbol{\theta}^{(i,j)},\boldsymbol{\beta}^{(i,j)},\boldsymbol{\epsilon}^{(i,j)}) = \mathrm{softmax}((\boldsymbol{\theta}^{(i,j)} + \boldsymbol{\beta}^{(i,j)}\boldsymbol{\epsilon}^{(i,j)})/\tau).
    \end{split}
\end{equation}
We illustrate some random samples of $\boldsymbol{\alpha}^{(i,j)}$ in Appendix~\ref{app:visual-sample} to show how $\boldsymbol{\beta}^{(i,j)}$ sharpens the distribution. 
The corresponding log probability density is ({detailed in Appendix~\ref{app:derive}}):
\vspace{-0.2cm}
\begin{equation}
\vspace{-0.1cm}
    \label{eq:logpdf}
    \small
    \Scale[0.85]{
    \begin{aligned}
        \log q(\boldsymbol{\alpha}^{(i,j)}|\boldsymbol{\theta}^{(i,j)}, \boldsymbol{\beta}^{(i,j)}) &= \log((K-1)!) + (K-1)\log\tau  - (K-1)\log\boldsymbol{\beta}^{(i,j)}-\sum_{k=1}^K\log\boldsymbol{\alpha}_k^{(i,j)}\\
        & + \sum_{k=1}^K\left(\frac{\boldsymbol{\theta}^{(i,j)}_k-\tau\log\boldsymbol{\alpha}^{(i,j)}_k}{\boldsymbol{\beta}^{(i,j)}}\right) - K\cdot\overset{K}{\underset{k=1}{\log\sum\exp}}\left(\frac{\boldsymbol{\theta}^{(i,j)}_k-\tau\log\boldsymbol{\alpha}^{(i,j)}_k}{\boldsymbol{\beta}^{(i,j)}}\right).
    \end{aligned}
    }
\end{equation}

With this, we can trivially derive the gradients of $\mathcal{L}(\boldsymbol{\theta}, \vw)$ \emph{w.r.t.} $\{\boldsymbol{\theta}, \vw\}$:
\begin{equation}
    \label{eq:gradient}
    \small
    \begin{split}
    \small
    \nabla_{\boldsymbol{\theta}} \mathcal{L}(\boldsymbol{\theta}, \vw) 
    = \mathbb{E}_{\boldsymbol{\epsilon}}[-\nabla_{\boldsymbol{\theta}} \log p(\mathcal{D}|g(\boldsymbol{\theta},\boldsymbol{\epsilon}), \vw) + \nabla_{\boldsymbol{\theta}} \log q(g(\boldsymbol{\theta},\boldsymbol{\epsilon})|\boldsymbol{\theta}) - \nabla_{\boldsymbol{\theta}} \log p(g(\boldsymbol{\theta},\boldsymbol{\epsilon}))], 
    \end{split}
\end{equation}
\begin{equation}
    \label{eq:gradientw}
    \small
    \nabla_{\vw} \mathcal{L}(\boldsymbol{\theta}, \vw) = \mathbb{E}_{\boldsymbol{\epsilon}}[-\nabla_{\vw} \log p(\mathcal{D}|g(\boldsymbol{\theta},\boldsymbol{\epsilon}), \vw)]+2\gamma\vw.
\end{equation}
In practice, we approximate the expectation in Eq.~(\ref{eq:gradient}) and~(\ref{eq:gradientw}) with $T$ Monte Carlo (MC) samples, and update $\boldsymbol{\alpha}$ and $\vw$ simultaneously. 

After training, let $\boldsymbol{\theta}^*$ and $\boldsymbol{w}^*$ denote the converged parameters, then we estimate the predictive distribution of unseen data by $
    p(y|x_{new},\boldsymbol{w}^*) = \mathbb{E}_{ q(\boldsymbol{\alpha}|\boldsymbol{\theta}^*)}[p(y|x_{new}, \boldsymbol{\alpha},\boldsymbol{w}^*)],
$
which is also known as \emph{Bayes ensemble}. For a tractable estimation, we use MC samples to approximate the expectation.
Note that DBSN assembles the predictions from networks whose structures are randomly sampled and weights are shared, which is in sharp difference from classic BNNs.

\vspace{-0.2cm}
\section{Related Work}
\vspace{-0.2cm}
Learning flexible Bayesian models has long been the goal of the community~\citep{mackay1992practical,neal1995bayesian,graves2011practical,blundell2015weight,balan2015bayesian,wang2016towards,gal2016dropout,kendall2017uncertainties,sun2017learning,louizos2017multiplicative,lakshminarayanan2017simple,zhang2018noisy,bae2018eigenvalue,khan2018fast}. 
In the spirit of structure learning, \citet{dikov2019bayesian} propose a unified Bayesian framework to infer the posterior of both the network weights and the structure, but the structure considered by them, \emph{i.e.}, layer size and network depth, 
is essentially impractical for complicated deep models. Instead, we inherit the structure representation of NAS to achieve more effective Bayesian structure learning for deep networks.

Neural architecture search (NAS)~\citep{zoph2016neural,zoph2018learning,pham2018efficient,real2019regularized}, especially the differentiable NAS~\citep{liu2018darts,xie2018snas,cai2018proxylessnas,wu2019fbnet}, has drawn tremendous attention recently. 
However, existing differentiable NAS methods work in a meta-learning way~\citep{finn2017model}, and need to re-train another network with the pruned compact structure after search. In contrast, DBSN unifies the learning of weights and structure in one stage, alleviating the structure mismatch and the inefficiency issues. 

\vspace{-0.2cm}
\section{Experiments}
\vspace{-0.2cm}
We evaluate DBSN by concerning two aspects: predictive performance and uncertainty estimation.

\textbf{Setup.} 
We set {\small $B=7$}, {\small $T=4$}, and {\small $\tau=\max(3\times\exp(-0.000015\mathrm{t}), 1)$} with $t$ as the global training step. 
We train DBSN for 100 epochs with batch size 64, which takes one day on 4 GTX 1080-Tis.
During test, we use \emph{100 MC samples} for Bayes ensemble. 
More details are given in Appendix~\ref{app:setup}. 

\textbf{Baselines.} 
We have implemented comparable baselines including {\emph{MAP}}, {\emph{MAP-fixed $\boldsymbol{\alpha}$}}, {\emph{MC dropout}}, {\emph{BBB}} (Bayes by Backprop), {\emph{FBN}} (fully Bayesian network), and {\emph{NEK-FAC}} (noisy EK-FAC), whose details are deferred to Appendix~\ref{app:baseline}.

\vspace{-0.2cm}
\subsection{Classification on CIFAR-10 and CIFAR-100}
\vspace{-0.1cm}
\label{sec:cl}
We repeat every classification experiment for 3 times and report the averaged error rate and standard deviation in Table~\ref{table:clf_abl}. 
As expected, DBSN is significantly more effective than the baselines.
There results confirm the supremacy of Bayesian treatment of structure over that of weights.

\begin{wraptable}{r}{0.5\textwidth}
  \caption{\footnotesize Comparison on classification error rate.}
  \vspace{-0.2cm}
  \label{table:clf_abl}
  \centering
  \footnotesize
  \setlength\tabcolsep{5.5pt}
  \begin{tabular*}{0.5\textwidth}{c|cc}
    \toprule    
    \textbf{Method}  &\textbf{CIFAR-10 (\%)}&\textbf{CIFAR-100 (\%)}\\
    \hline
    {DBSN} &  $\mathbf{4.98}\pm0.24$ & $\mathbf{22.50}\pm0.26$\\
    \hline
    {MAP} & $5.79\pm0.34$ & $24.19\pm0.17$\\
    {MAP-fixed $\boldsymbol{\alpha}$}  & $5.66 \pm 0.24$ & $24.27\pm0.15$\\
    {MC dropout} & $5.83\pm 0.19$ & $23.67\pm0.28$\\
    \red{{BBB}} & $9.85\pm0.42$ & $30.98\pm0.36$\\
    \red{{FBN}} & $9.57\pm0.55$ & $31.39\pm0.06$\\
    {NEK-FAC}  & $7.43$ & $37.47$\\
    \bottomrule
  \end{tabular*}
  \vspace{-0.2cm}
\end{wraptable} 
We defer the comparison with more baselines to Appendix~\ref{app:comp-dnns} and \ref{app:vogn}.
Besides, we provide a refinement of DBSN by explicitly modeling the dependency of weights on structure:
we sample 10 structures from the learned structure posterior of DBSN, and train 10 corresponding networks individually, then uniformly assemble them to predict. 
The error rate on CIFAR-10 is 3.96\%, slightly surpassing DBSN (4.98\%). 
As a baseline, we randomly select one from the 10 structures and train 10 networks all with that structure, whose ensemble yields 4.33\% error rate. This comparison further confirms the benefits of structure stochasticity for Bayes ensemble.

\begin{table}[t]
  \vspace{-0.1cm}
  \caption{\footnotesize  Comparison on model calibration in terms of the Expected Calibration Error (ECE). Smaller is better. 
  }
  \vspace{-0.3cm}
  \label{table:ece}
  \small
  \setlength\tabcolsep{7pt}
  \centering
  \begin{tabular*}{\textwidth}{c|c|c|c|c|c|c|c}
    \toprule
    {\textbf{Method}} & {DBSN} & {MAP} & { MAP-fixed $\boldsymbol{\alpha}$}  & {MC dropout} & \red{{BBB}} & \red{{FBN}} & {NEK-FAC}\\
    \hline
    \textbf{CIFAR-10} & $\mathbf{0.0109}$& $0.0339$ & $0.0327$   & $0.0150$& $0.0745$ & $0.0966$ & $0.0434$\\
    \textbf{CIFAR-100}  & $\mathbf{0.0599}$& $0.1240$  & $0.1259$  & $0.0617$ & $0.0700$ & $0.1091$& $0.1665$\\
    
    \bottomrule
  \end{tabular*}
  \vspace{-0.1cm}
\end{table}

We then assess if DBSN can yield calibrated predictive uncertainty by quantifying model calibration. 
For this purpose, we compute the Expected Calibration Error (ECE)~\citep{guo2017calibration} and depict the comparison in Table~\ref{table:ece}.
We also plot some reliability diagrams~\citep{guo2017calibration} in Appendix~\ref{app:cali} to provide a direct explanation of calibration.
DBSN outperforms all the baselines, revealing its practical value. 
As reference, we also train a DenseNet-40-12 (1M) and a DenseNet-BC-100-12 (0.8M) on CIFAR-10. 
The two models yield 0.0338 and 0.0290 ECEs (with 5.55\% and 4.89\% error rates), highlighting that DBSN (0.0109 ECE and 4.98\% error rate) conjoins good accuracy and calibration.
We defer more comparisons on the quality of uncertainty estimates to Appendix~\ref{app:pred-unc}.

\vspace{-0.3cm}
\subsection{Semantic Segmentation on CamVid}
\begin{figure}[t!]
\centering
\begin{subfigure}{0.24\textwidth}
  \centering
  \includegraphics[width=\linewidth]{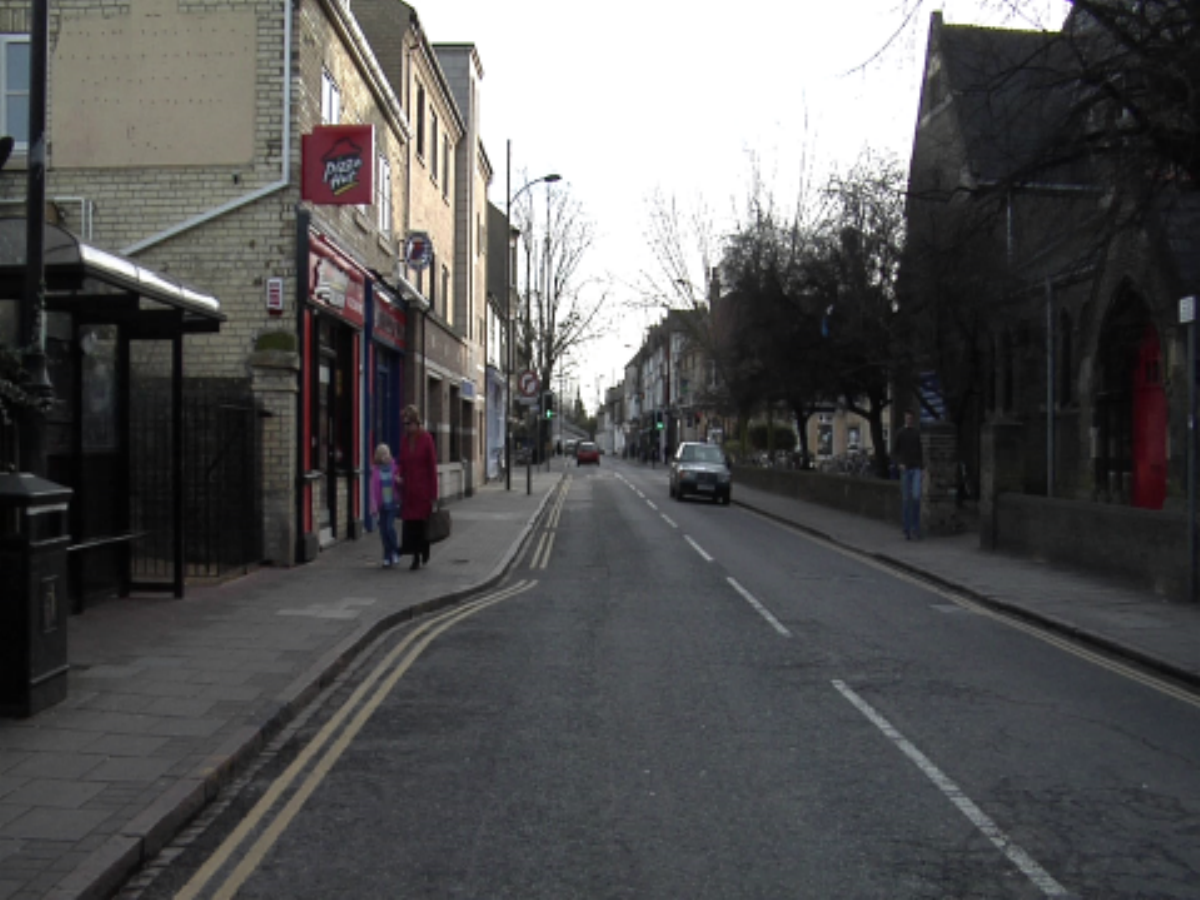}
  \label{fig:sr-1}
\end{subfigure}
\begin{subfigure}{0.24\textwidth}
  \centering
  \includegraphics[width=\linewidth]{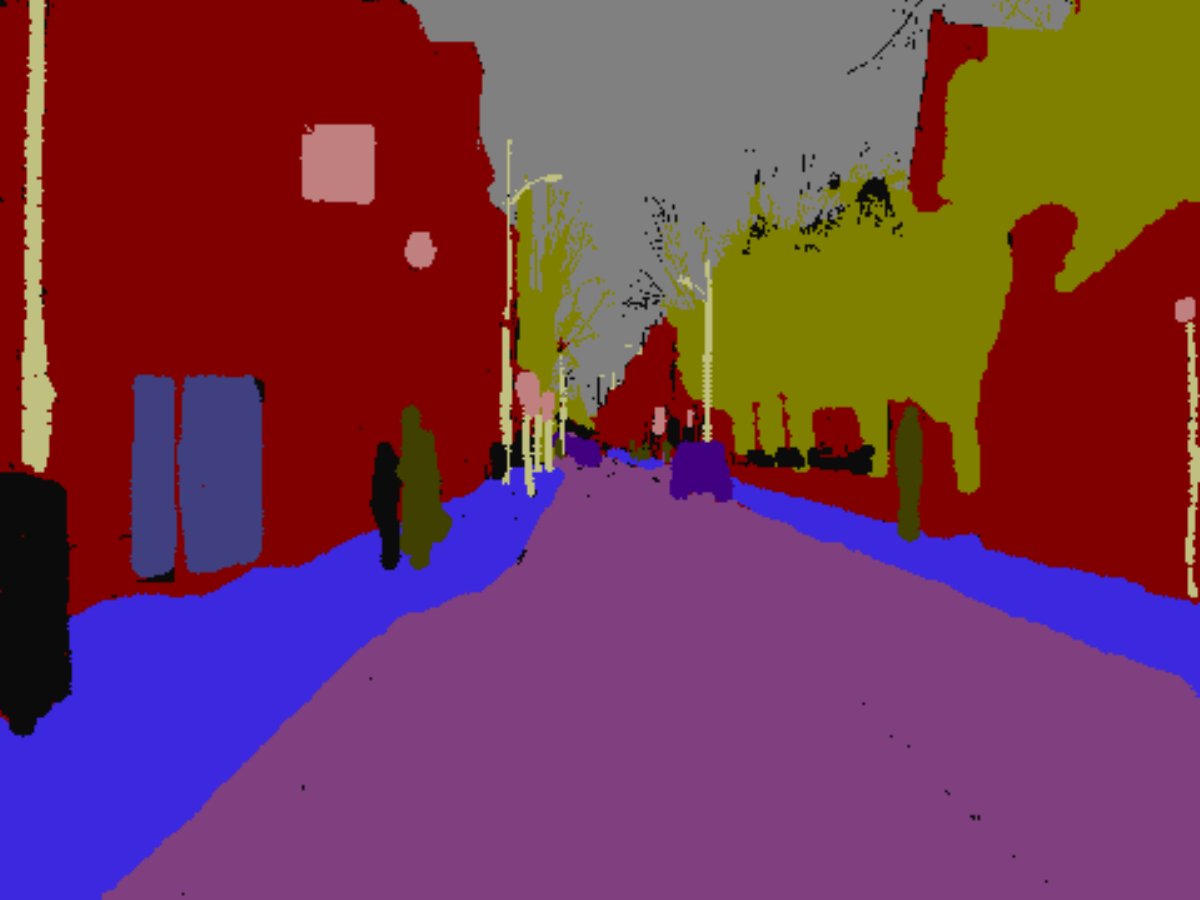}
  \label{fig:sr-2}
\end{subfigure}
\begin{subfigure}{0.24\textwidth}
  \centering
  \includegraphics[width=\linewidth]{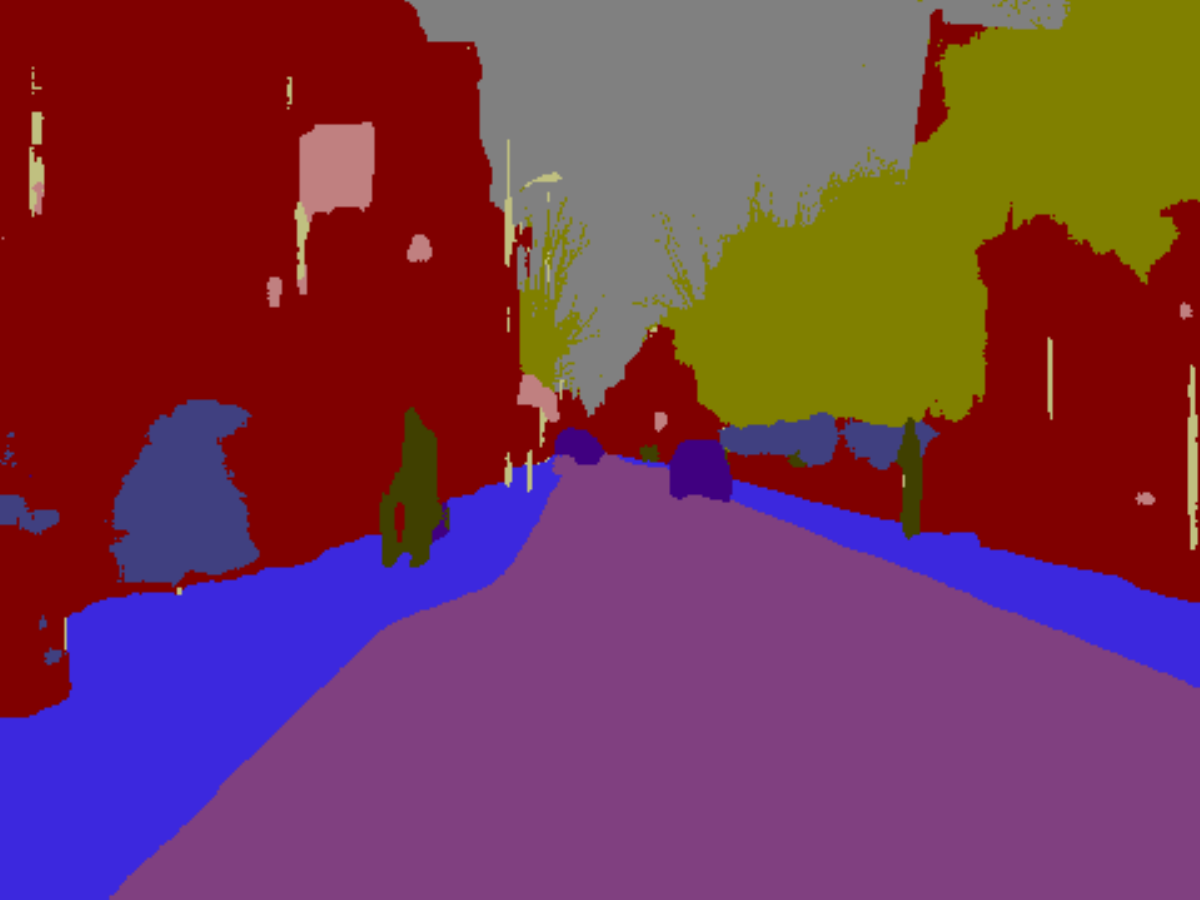}
  \label{fig:sr-3}
\end{subfigure}
\begin{subfigure}{0.24\textwidth}
  \centering
  \includegraphics[width=\linewidth]{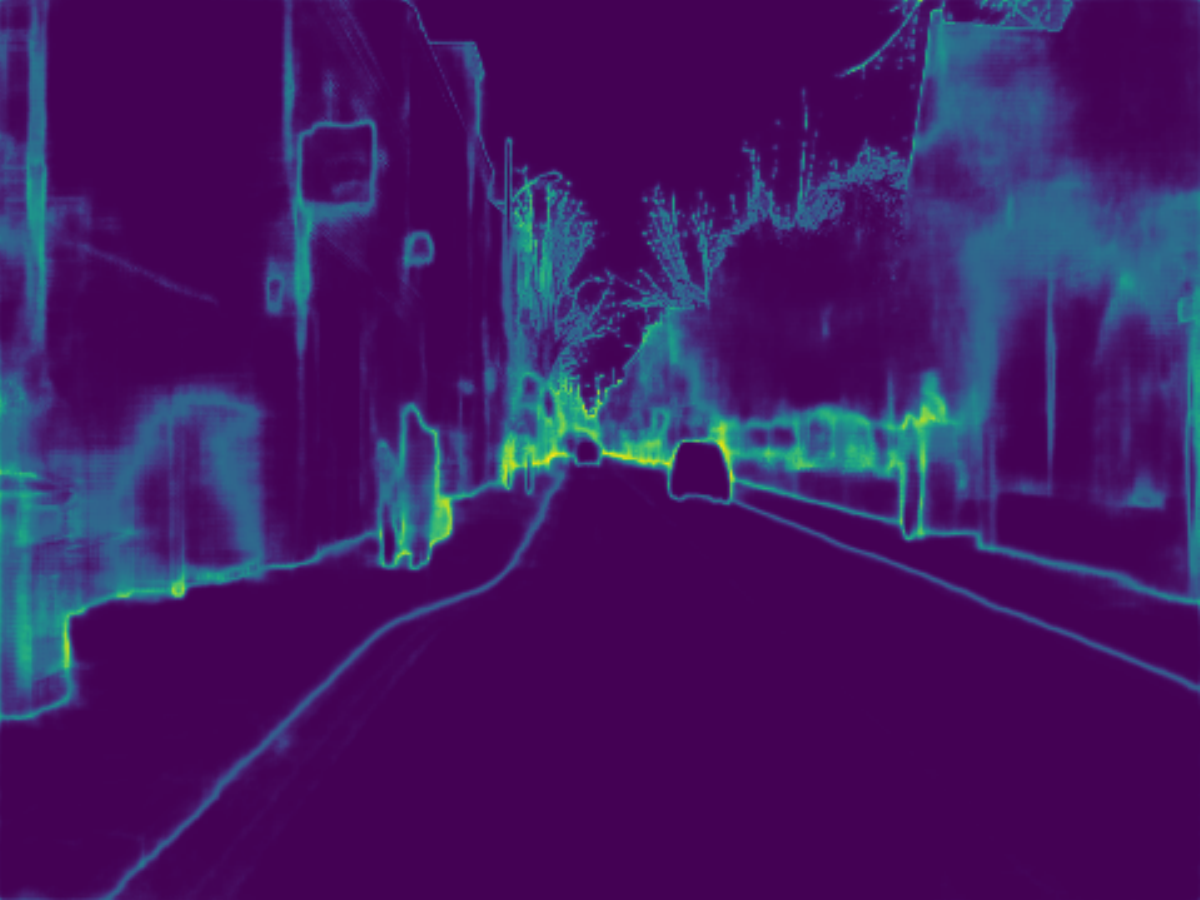}
  \label{fig:sr-4}
\end{subfigure}
\vspace{-0.6cm}
\caption{\footnotesize  Visualization of the segmentation results of DBSN on CamVid. From left to right: original image, ground-truth segmentation, the estimated segmentation, and pixel-wise predictive uncertainty. }\vspace{-0.cm}
\label{fig:seg-results}
\vspace{-0.5cm}
\end{figure}

\vspace{-0.25cm}
\label{sec:seg}
To show DBSN is readily applicable to diverse scenarios, we extend DBSN to the challenging semantic segmentation task on CamVid~\citep{brostow2008segmentation}.
Our implementation is based on the brief FC-DenseNet framework~\citep{jegou2017one}. 
We present the results in Table~\ref{table:seg} and Figure~\ref{fig:seg-results}.
It is evident that DBSN surpasses the competing FC-DenseNet67 by a large margin while using fewer parameters. 
DBSN also demonstrates significantly better performance than the classic Bayesian SegNet which adopts MC dropout for uncertainty estimation. 
In Figure~\ref{fig:seg-results}, it is also worth noting that the uncertainty produced by DBSN is interpretable: the edges of the objects and the regions which contain overlapping have substantially higher uncertainty than the other parts.
These results validate the potential of DBSN to be applied into diverse, challenging tasks.

\begin{table}[t]
  \vspace{-0.1cm}
  \centering
  \caption{\footnotesize Comparison on segmentation performance. * indicates results from our implementation. (CamVid)}
  \vspace{-0.3cm}
  \label{table:seg}
  \setlength\tabcolsep{7pt}
  \footnotesize 
  \begin{tabular*}{\textwidth}{c|c|c|c|c}
    \toprule
    \textbf{Method} & \textbf{Pretrained} & \textbf{Params (M)} & \textbf{Mean IoU} & \textbf{Global accuracy}\\
    \hline
    {SegNet}~\citep{badrinarayanan2015segnet} & \cmark & $29.5$ & $46.4$ & $62.5$\\
    {Bayesian SegNet}~\citep{kendall2015bayesian} & \cmark & $29.5$ & $63.1$ & $86.9$\\
    \hline
    {FC-DenseNet67}~\citep{jegou2017one} & \xmark & $3.5$ & $63.1$* & $90.4$*\\
    {DBSN} & \xmark & $3.3$ & $\bm{65.4}$ & $\bm{91.4}$\\
    \bottomrule
  \end{tabular*}
\end{table}

\vspace{-0.2cm}
\section{Conclusion}
\vspace{-0.25cm}
In this work, we have introduced a novel Bayesian structure learning approach for DNNs. The proposed DBSN draws the inspiration from NAS and models the network structure as Bayesian variables. Stochastic variational inference is employed to jointly learn the weights and the distribution of the structure. 
We evaluate DBSN in diverse scenarios and witness good results.

\bibliography{iclr2021_conference}
\bibliographystyle{iclr2021_conference}

\appendix

\section{Details of the Sharpened Concrete Distribution}
\subsection{Visualization of the Sample}
\label{app:visual-sample}

\begin{figure}
    \centering
    \includegraphics[width=0.8\linewidth]{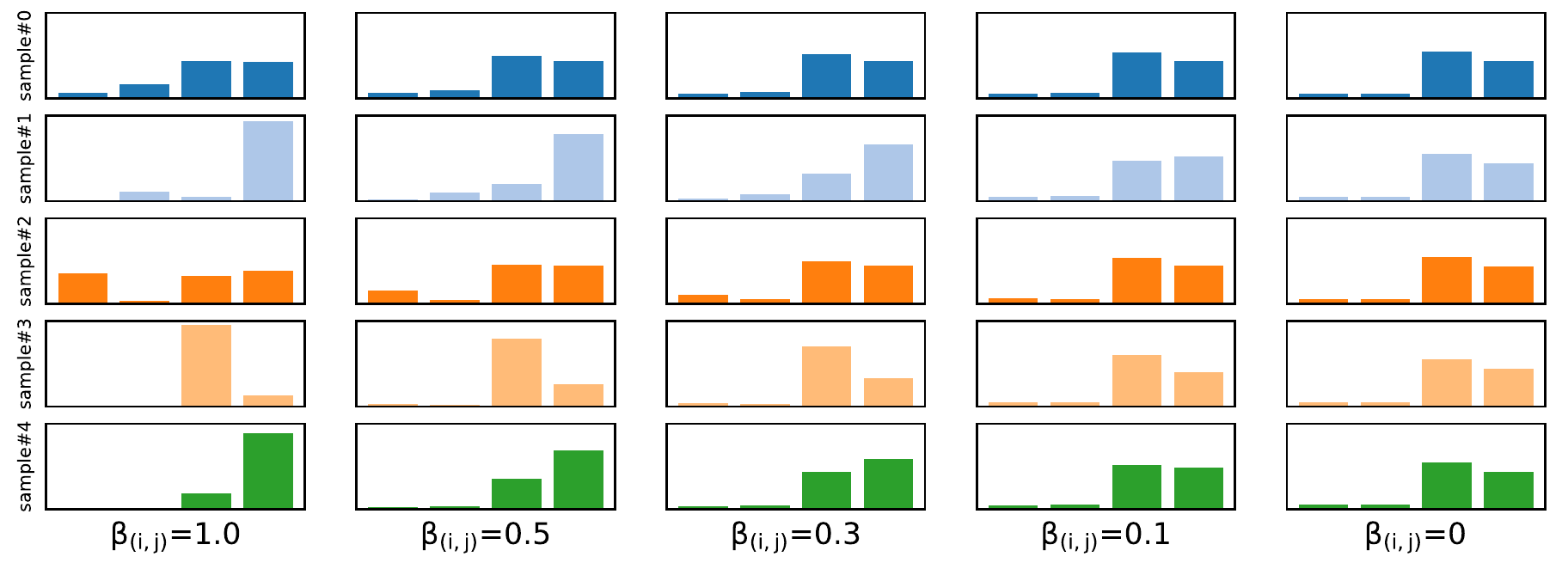}
\vspace{-1.7ex}
\caption{\footnotesize Each column includes 5 samples from Eq.~(\ref{eq:concrete}) with the same $\boldsymbol{\beta}^{(i,j)}$. Samples in every row share the same $\boldsymbol{\epsilon}^{(i,j)}$. The base class probabilities of each sample are $\mathrm{softmax}(\boldsymbol{\theta}^{(i,j)}) = [0.05, 0.05, 0.5, 0.4]$, a common pattern in the learned structure distribution.}
\label{fig:con}
\end{figure}

As shown in Figure~\ref{fig:con}, sliding $\boldsymbol{\beta}^{(i,j)}$ from 1 to 0 
gradually decreases the diversity of the sampled structures, leading to a colder and colder posterior. 
When $\boldsymbol{\beta}^{(i,j)}=0$, the estimation on the structure degenerates to MAP, making us lose the benefits of Bayesian principles.
Thus, we decide to sharpen the posterior with a pre-defined schedule to control the posterior dynamics more reliably.
For example, in practice, we gradually decay $\boldsymbol{\beta}^{(i,j)}$ from 1 to 0.5 along with the training process.

\subsection{Derivation of the Log Probability Density}
\label{app:derive}

For clear expressions, We simply denote $\boldsymbol{\alpha}^{(i,j)}$, $\boldsymbol{\theta}^{(i,j)}$, $\boldsymbol{\beta}^{(i,j)}$ and $\boldsymbol{\epsilon}^{(i,j)}$ as $\boldsymbol{\alpha}$, $\boldsymbol{\theta}$, $\beta$ and $\boldsymbol{\epsilon}$, respectively. Let $\vp=\mathrm{softmax}(\boldsymbol{\theta})$. Consider
\begin{equation*}
    \begin{split}
    \boldsymbol{\alpha}_k &= \frac{\exp((\boldsymbol{\theta}_k+\beta\boldsymbol{\epsilon}_k)/\tau)}{\sum_{i=1}^K\exp((\boldsymbol{\theta}^i+\beta\boldsymbol{\epsilon}^i)/\tau)} \\&= \frac{\exp((\log\vp_k+\beta\boldsymbol{\epsilon}_k)/\tau)}{\sum_{i=1}^K\exp((\log\vp^i+\beta\boldsymbol{\epsilon}^i)/\tau)}.
    \end{split}
\end{equation*}
Let $\vz_k=\log\vp_k+\beta\boldsymbol{\epsilon}_k=\log\vp_k-\beta\log(-\log(\vu_k))$, where $\vu_k \sim \mathcal{U}(0,1)\;\mathrm{i.i.d.}$. It has density
\begin{equation*}
    \frac{1}{\beta}\vp_k^{1/\beta}\exp(-\frac{\vz_k}{\beta})\exp(-\vp_k^{1/\beta}\exp(-\frac{\vz_k}{\beta})).
\end{equation*}
We denote $c=\sum_{i=1}^K\exp(\vz_i/\tau)$, then $\boldsymbol{\alpha}_k=\exp(\vz_k/\tau)/c$. Consider this transformation:
\begin{equation*}
    F(\vz_1,\dots,\vz_K)=(\boldsymbol{\alpha}_1,\dots,\boldsymbol{\alpha}_{K-1},c),
\end{equation*}
which has the following inverse transformation: \begin{equation*}
\begin{gathered}
F^{-1}(\boldsymbol{\alpha}_1,\dots,\boldsymbol{\alpha}_{K-1},c)=(\tau(\log\boldsymbol{\alpha}_1 + \log c),\dots,\\ \tau(\log\boldsymbol{\alpha}_K + \log c)),
\end{gathered}
\end{equation*}
whose Jacobian has the determinant (refer to the derivation of the concrete distribution~\citep{maddison2016concrete}):
\begin{equation*}
    \frac{\tau^K}{c\prod_{i=1}^K\boldsymbol{\alpha}_i}.
\end{equation*}

Multiply this with the density of $\vz$, we get the density
\begin{equation*} 
\begin{gathered}
    \frac{\tau^K}{c\prod_{i=1}^K\boldsymbol{\alpha}_i} \prod_{i=1}^K[\frac{1}{\beta}\vp_i^{1/\beta}\exp(-\frac{\tau(\log\boldsymbol{\alpha}_i + \log c)}{\beta})  \\ \cdot \exp(-\vp_i^{1/\beta}\exp(-\frac{\tau(\log\boldsymbol{\alpha}_i + \log c)}{\beta}))].
\end{gathered}
\end{equation*}

Let $r=\log c$, then apply the change of variables formula, we easily obtain the density:
\begin{equation*} 
\begin{gathered}
\frac{\tau^K\prod_{i=1}^K\vp_i^{1/\beta}}{\beta^K\prod_{i=1}^K \boldsymbol{\alpha}_i^{(1+\tau/\beta)}} \exp(-\frac{K\tau r}{\beta})\\
\cdot\exp(-\sum_{i=1}^K(\vp_i \boldsymbol{\alpha}_i^{-\tau})^{1/\beta}\exp(-\frac{\tau r}{\beta})).
\end{gathered}
\end{equation*}

We use $\gamma$ to substitute $\log\sum_{i=1}^K(\vp_i \boldsymbol{\alpha}_i^{-\tau})^{1/\beta}$, then get:
\begin{equation*} 
\begin{gathered}
    \frac{\tau^K\prod_{i=1}^K\vp_i^{1/\beta}}{\exp(\gamma)\beta^K\prod_{i=1}^K \boldsymbol{\alpha}_i^{(1+\tau/\beta)}}\exp(\gamma-\frac{K\tau r}{\beta})\\ \cdot\exp(-\exp(\gamma-\frac{\tau r}{\beta})).
\end{gathered}
\end{equation*}

Naturally, we can integrate out $r$, and get:
\begin{equation*}
\begin{split}
    &\frac{\tau^K\prod_{i=1}^K\vp_i^{1/\beta}}{\exp(\gamma)\beta^K\prod_{i=1}^K \boldsymbol{\alpha}_i^{(1+\tau/\beta)}}\left[\frac{\beta}{\tau}\exp(\gamma-K\gamma)\Gamma(K)\right]\\
    =&\frac{\tau^{K-1}\prod_{i=1}^K\vp_i^{1/\beta}}{\beta^{K-1}\prod_{i=1}^K \boldsymbol{\alpha}_i^{(1+\tau/\beta)}}\exp(-K\gamma)\Gamma(K)\\
    =&\frac{((K-1)!)\tau^{K-1}}{\beta^{K-1}\prod_{i=1}^K \boldsymbol{\alpha}_i} \times \frac{\prod_{i=1}^K(\vp_i \boldsymbol{\alpha}_i^{-\tau})^{1/\beta}}{(\sum_{i=1}^K(\vp_i \boldsymbol{\alpha}_i^{-\tau})^{1/\beta})^K}.
\end{split}
\end{equation*}
Then, the log density is:
\begin{equation*}
\begin{split}
    \log&((K-1)!) + (K-1)\log\frac{\tau}{\beta} - \sum_{i=1}^K\log\boldsymbol{\alpha}_i \\&+ \sum_{i=1}^K\frac{\log\vp_i-\tau\log\boldsymbol{\alpha}_i}{\beta} - K \cdot\overset{K}{\underset{i=1}{\mathrm{L\Sigma E}}}\frac{\log\vp_i-\tau\log\boldsymbol{\alpha}_i}{\beta}\\
\end{split}
\end{equation*}
\begin{equation*}
\begin{split}
    =\log&((K-1)!) + (K-1)\log\frac{\tau}{\beta} - \sum_{i=1}^K\log\boldsymbol{\alpha}_i \\&+ \sum_{i=1}^K\frac{\boldsymbol{\theta}_i-\tau\log\boldsymbol{\alpha}_i}{\beta} - K \cdot\overset{K}{\underset{i=1}{\mathrm{L\Sigma E}}}\frac{\boldsymbol{\theta}_i-\tau\log\boldsymbol{\alpha}_i}{\beta},
\end{split}
\end{equation*}
which is equal to Eq.~(\ref{eq:logpdf}).

\section{Detailed Experiment settings}
\subsection{Network and Training Details}
\label{app:setup}

\textbf{Support reduction.} Given the combinatorial nature of the structure, the support of the structure posterior is exponentially large, perhaps leading to undesirable over-regularization on weights. 
Furthermore, as can be seen from Eq.~(\ref{eq:arch}), only the weights of the operations activated by current structure sample are updated in one training step, thus we may require more (up to $K\times$) training steps to drive the model to converge.
Therefore, we propose to reduce the support of the structure posterior, by removing inessential operations from the operation set of DARTS~\citep{liu2018darts}, including all the 5$\times$5 convolutions that can be replaced by stacked 3$\times$3 convolutions and all the pooling layers which are mainly used for the downsampling module, and defining the structure as the masks on the others, \emph{i.e.}, 3$\times$3 separable convolutions, 3$\times$3 dilated separable convolutions, identity and \emph{zero}. 

\textbf{Activating the expressivity of BN.} 
As suggested by DARTS~\citep{liu2018darts}, we use ReLU-Conv-BN style operations in DBSN. 
However, the learnable affine transformations in batch normalizations (BNs)~\citep{ioffe2015batch} are required to be disabled to prevent the scaling effect of the structure mask being cancelled out.
As a consequence, the expressivity of the model is discounted.
To address this issue and in turn to enhance the data fitting of DBSN, we propose to place a complete batch normalization in the front of the next operation. Namely, we adopt the BN-ReLU-Conv-BN style operations, where the first BN has learnable affine parameters while the second one does not.

\textbf{More details of the classification experiments.} 
The candidate operations all have 16 output channels. 
We concatenate all the intermediate nodes along with the input to get the cell's output. 
The whole network is composed of 12 cells and 2 downsampling modules which have a channel compression factor of 0.4 and are located at the 1/3 and 2/3 depth. 
We constrain downsampling modules to be the typical BN-ReLU-Conv-Pooling operations, to ease structure learning. 
We employ a 3$\times$3 convolution before the first cell and put a global average pooling followed by a fully connected (FC) layer after the last cell. 
We set $\tau=\max(3\times\exp(-0.000015\mathrm{t}), 1)$ where $t$ is the global training step. 
We initialize $\vw$ and $\boldsymbol{\theta}$ following \cite{he2015delving} and \cite{liu2018darts}, respectively. 
A momentum SGD with initial learning rate 0.1 (divided by 10 at 50\% and 75\% of the training procedure following~\cite{huang2017densely}), momentum $0.9$ and weight decay $10^{-4}$ is used to train the weights $\vw$. We clip the gradient norm $||\nabla_{\vw}||$ at $5$. 
An Adam optimizer with learning rate $3\times 10^{-4}$, momentum ($0.5$, $0.999$) is used to learn $\boldsymbol{\theta}$. We deploy the standard data augmentation scheme (mirroring/shifting) and normalize the data with the channel statistics. The whole training set is used for optimization.

\textbf{More details of the segmentation experiments.} We only replace the original dense blocks with structure-learnable cells, without introducing further advanced techniques from the semantic segmentation community, to figure out the performance gain only from the proposed structure learning approach.
For the setup, we set $B=5$ (same as the number of layers in every dense block of FC-DenseNet67) and $T=1$, and learn two cell structures for the downsampling path and upsampling path, respectively. 
We use a momentum SGD with initial learning rate 0.01 (which decays linearly after 350 epochs), momentum 0.9 and weight decay $10^{-4}$ instead of the original RMSprop for better results. 
The other settings follow \citet{jegou2017one} and the classification experiments.

\subsection{Elaborated Baselines}
\label{app:baseline}
We have implemented a wide range of baselines, most of which have learnable structures to make fair comparison with DBSN, including: 
$(\RN{1})$ \textbf{\emph{MAP}}: a variant of \emph{DBSN} with both point-estimate weights and point-estimate structure under L2 penalty;
$(\RN{2})$ \textbf{\emph{MAP-fixed $\boldsymbol{\alpha}$}}, a variant of \emph{MAP} with fixed, uniform structure masks; 
$(\RN{3})$ \textbf{\emph{MC dropout}}: a variant of \emph{MAP} with dropout after every convolution (0.2 dropout rate);
$(\RN{4})$ \textbf{\emph{BBB}}: a variant of \emph{MAP} with weight uncertainty (and point-estimate structure), where the fully factorized Gaussians are deployed on weights and Bayes by Backprop~\citep{blundell2015weight} is used for inference;
$(\RN{5})$ \textbf{\emph{FBN}} (fully Bayesian network): a variant of \emph{DBSN} with weight uncertainty, where Bayes by Backprop is also adopted for inference;
$(\RN{6})$ \textbf{\emph{NEK-FAC}}: a \emph{VGG16} network (3.7M parameters) with weight uncertainty, where the noisy EK-FAC~\citep{bae2018eigenvalue} algorithm is used for inference (it is hard to implement this algorithm with learnable structure); 
$(\RN{7})$ \textbf{\emph{VOGN}}: a variant of \emph{DBSN} with weight uncertainty, where the Variational Online Gauss-Newton~\citep{khan2018fast} is employed to infer weight posterior (but due to its high inefficiency, we only provide its results in small model size in Appendix~\ref{app:vogn}).
Note that $(\RN{4})$-$(\RN{7})$ all locate in the variational BNN family.

\section{More Empirical Results}
\subsection{Comparison with Popular DNNs}
\label{app:comp-dnns}
We compare DBSN to popular DNNs in aspect of predictive performance on CIFAR-10 and CIFAR-100.
The results are reported in Table~\ref{table:clf}. 
As shown, DBSN is notably comparable with or even better than the popular DNN models. DBSN only presents modestly higher error rates than DenseNet-BC~\citep{huang2017densely}. The comparisons highlight the practical value of DBSN. 
\begin{table}[t]
  \caption{\footnotesize Comparison with popular DNNs in terms of the number of parameters and test error rate.}
  \vspace{-0.3cm}
  \label{table:clf}
  \centering \footnotesize
  \setlength\tabcolsep{10pt}
  \begin{tabular*}{\textwidth}{c|ccc}
    \toprule
    \textbf{Method}     & \textbf{Params (M)}  &\textbf{CIFAR-10 (\%)}&\textbf{CIFAR-100 (\%)}\\
    \hline
    {ResNet~\citep{he2016deep}} & $1.7$ & $6.61$ & -\\
    {ResNet (pre-activation)~\citep{he2016identity}} & $1.7$ & $5.46$ & $24.33$\\
    {DenseNet~\citep{huang2017densely}} & $1.0$ & $5.24$ & $24.42$\\
    {DenseNet-BC~\citep{huang2017densely}} & $0.8$ & { $\mathbf{4.51}$} & { $\mathbf{22.27}$}\\
    \hline
    {DBSN} & $1.0$ &  $\mathbf{4.98}\pm0.24$ & $\mathbf{22.50}\pm0.26$\\
    \bottomrule
  \end{tabular*}
\end{table}

\subsection{\red{Comparison with Competing Baselines using VOGN for Inferring Weight Posterior}}
\label{app:vogn}

\begin{table*}[t]
  \vspace{-0.cm}
  \caption{\red{\footnotesize Comparison between DBSN and competing baselines which deploy uncertainty on weights and adopt VOGN algorithm for variational inference on weights. Baseline \emph{VOGN} performs posterior inference on weights but performs MAP estimation on structure. Baseline \emph{VOGN-FBN} 
  performs posterior inference on both weights and structure, constructing a fully Bayesian network. (CIFAR-10)}}
  \label{table:vogn}
  \vspace{-0.3cm}
  \setlength\tabcolsep{15pt}
  \centering
  \footnotesize
  \begin{tabular*}{0.94\textwidth}{l|c|c|c}
    \toprule
    \textbf{} & \textbf{Training time (hours)} & \textbf{Test error rate (\%)} & \textbf{ECE}\\
    \hline
    \red{{DBSN}}& $1.2$ & $9.90$ & $0.0070$\\
    \emph{VOGN} & $13.0$ & $28.4$& $0.5391$\\
    \emph{VOGN-FBN}& $13.0$ & $30.5$& $0.5169$\\
    \bottomrule
  \end{tabular*}
\end{table*}

We realize the Bayes by Backprop~\citep{blundell2015weight} method used for inferring weight posterior in {BBB} and {FBN} baselines may be restrictive, resulting in such weakness. Therefore, we replace it with Variational Online Gauss-Newton (VOGN)~\citep{khan2018fast,osawa2019practical}, a most-recently proposed mean-field natural-gradient variational inference method. 
VOGN is known to work well with advanced techniques, e.g., momentum, batch normalisation, data augmentation. 
As claimed by \cite{osawa2019practical}, VOGN demonstrates comparable results to Adam. 
Our implementation is based on VOGN's official repository (\url{https://github.com/team-approx-bayes/dl-with-bayes}). 
With the original network size ($B=7$, 12 cells), the baselines trained with VOGN need more than one hour for one epoch, which is very slow. 
Thus we adopt smaller networks ($B=4$, 3 cells), which have almost 41K parameters, for the two baselines {BBB} and {FBN}. 
We also train a DBSN in the same setting.
The experiments are conducted on CIFAR-10 and the results are provided in Table~\ref{table:vogn}. {The predictive performance and uncertainty gaps between DBSN and the two baselines are very huge, which possibly results from the under-fitting of the high-dim weight distributions}. 
We think that our implementation is correct because our results are consistent with the original results in Table 1 of \cite{osawa2019practical} (VOGN has 75.48\% and 84.27\% validation accuracy even with even larger 2.5M AlexNet and 11.1M ResNet-18 architectures). 
Further, DBSN is {much more efficient} than them. 
These comparisons strongly reveal the benefits of modeling structure uncertainty over modeling weight uncertainty.

\vspace{-0.1cm}
\subsection{Comparison on Reliability Diagrams for Evaluating Model Calibration}
\begin{figure*}[t]
\centering
\begin{subfigure}{0.32\textwidth}
  \centering
  \includegraphics[width=\linewidth]{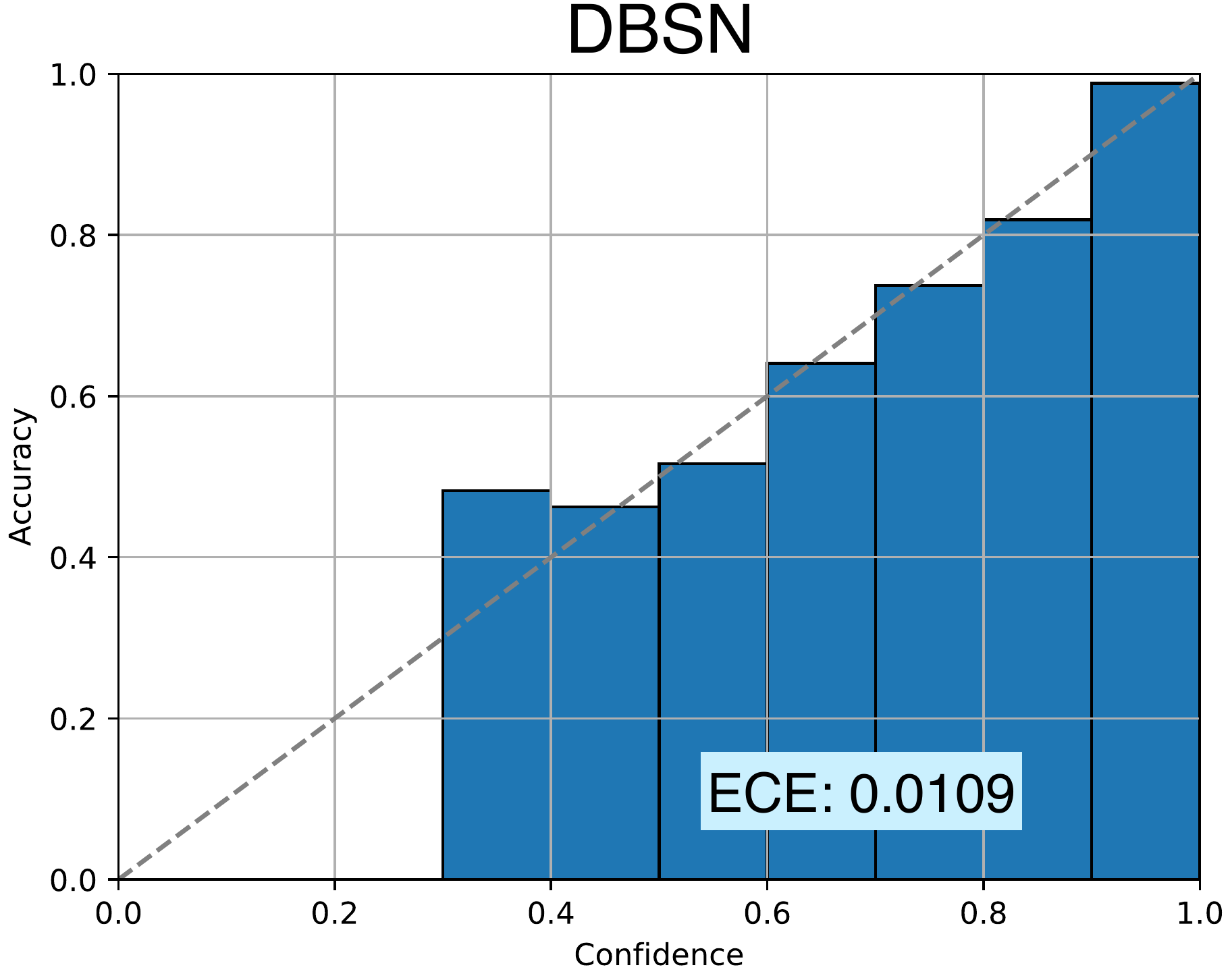}
\end{subfigure}
\begin{subfigure}{0.32\textwidth}
  \centering
  \includegraphics[width=\linewidth]{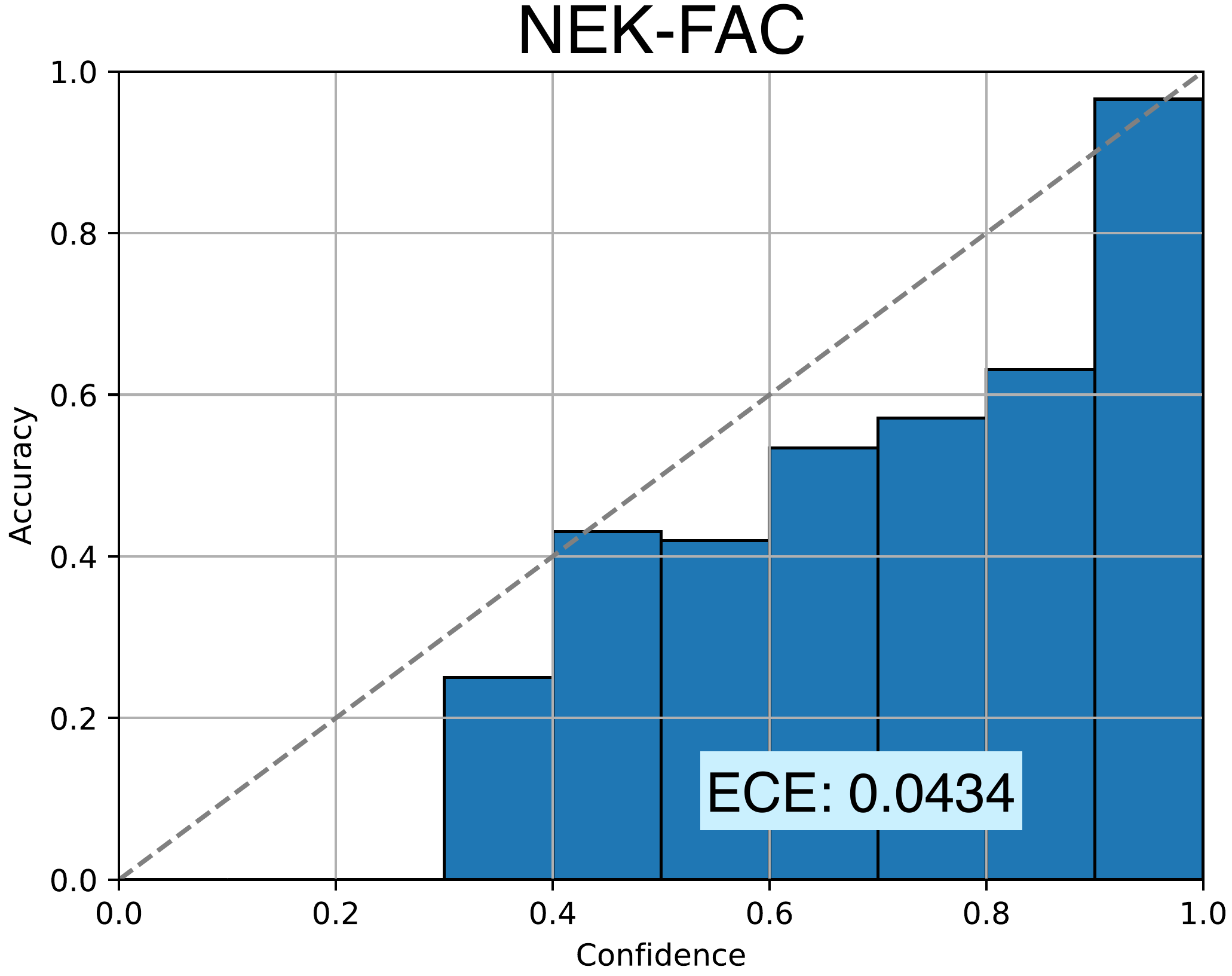}
\end{subfigure}
\begin{subfigure}{0.32\textwidth}
  \centering
  \includegraphics[width=\linewidth]{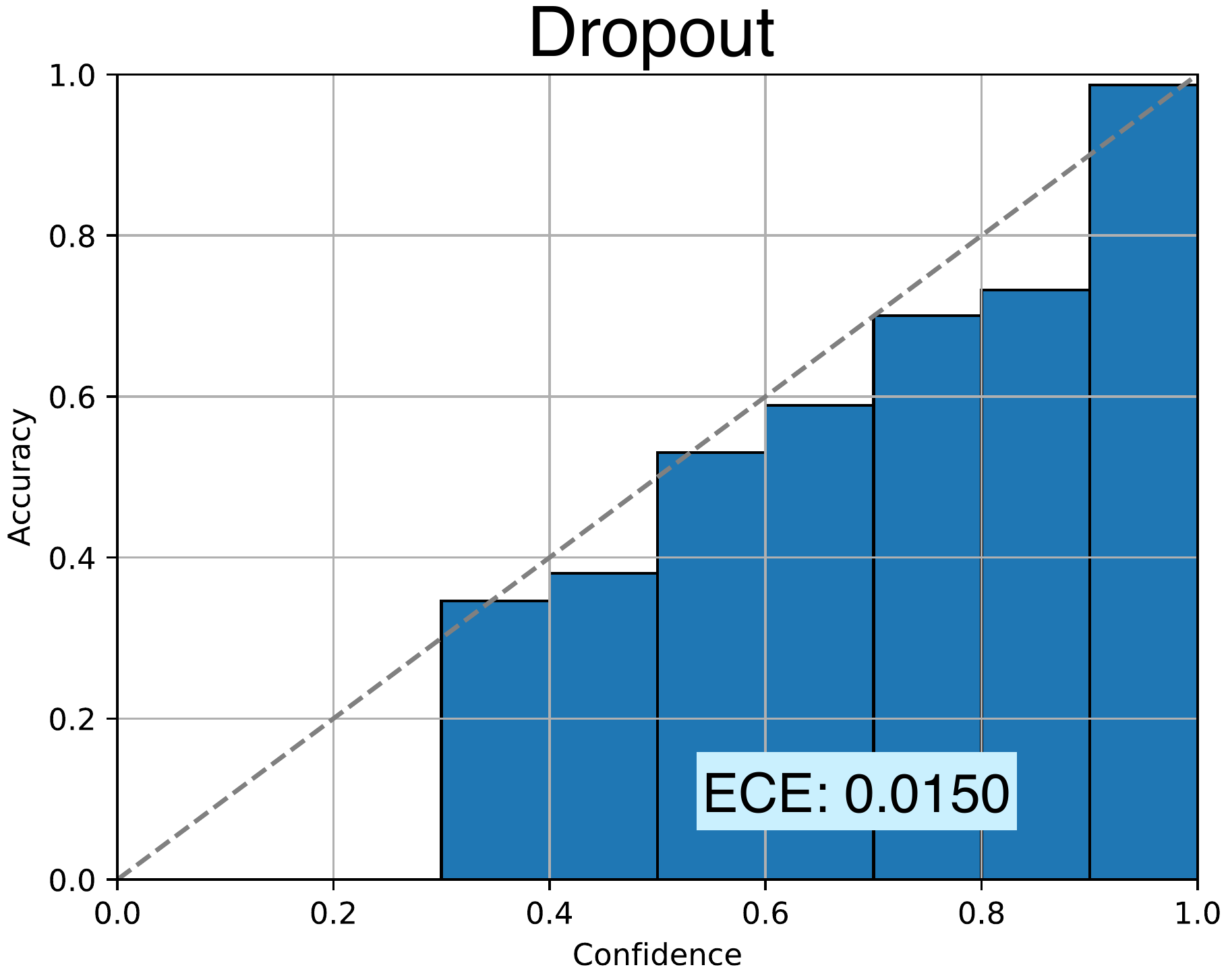}
\end{subfigure}
\begin{subfigure}{0.32\textwidth}
  \centering
  \includegraphics[width=\linewidth]{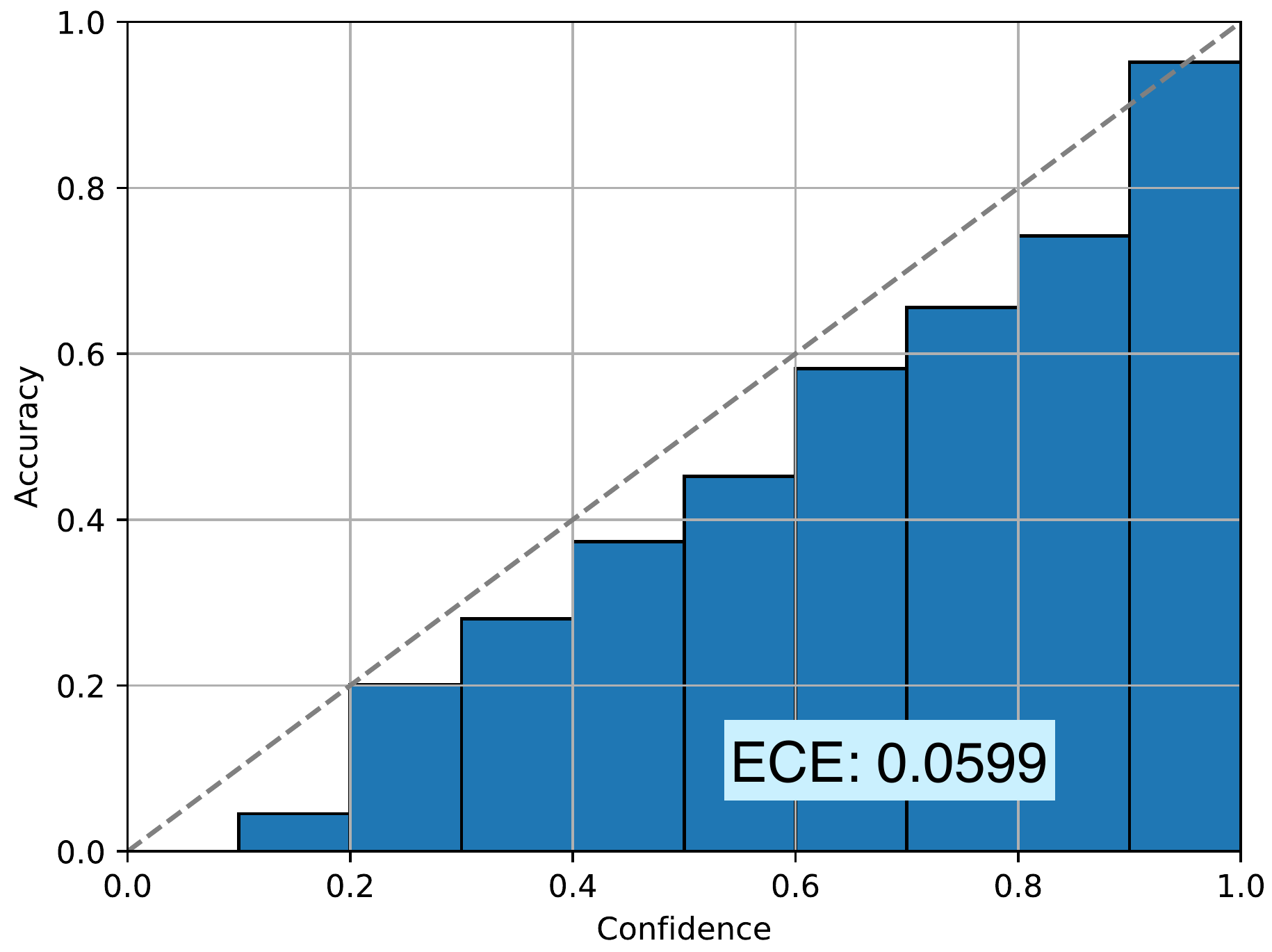}
\end{subfigure}
\begin{subfigure}{0.32\textwidth}
  \centering
  \includegraphics[width=\linewidth]{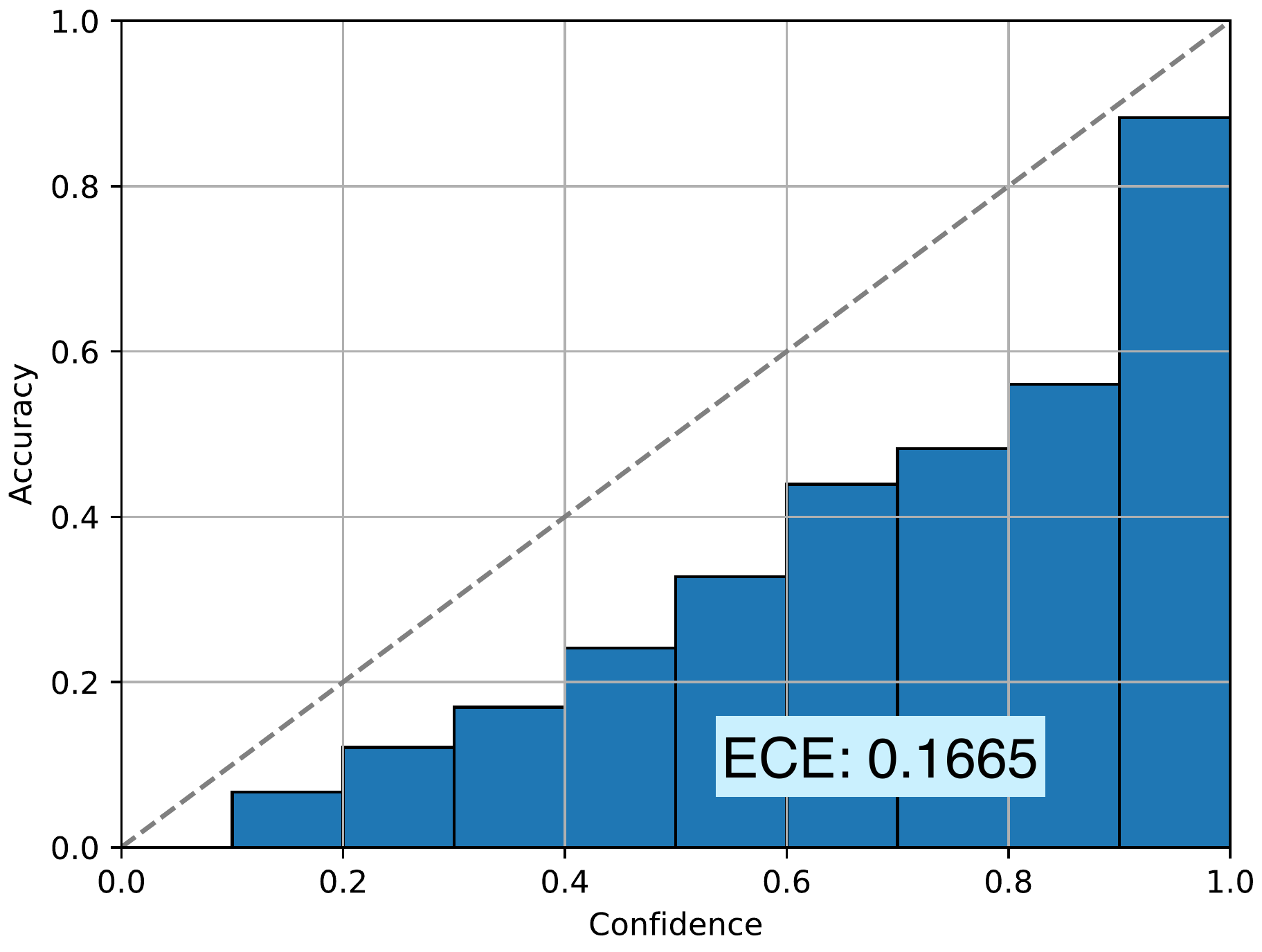}
\end{subfigure}
\begin{subfigure}{0.32\textwidth}
  \centering
  \includegraphics[width=\linewidth]{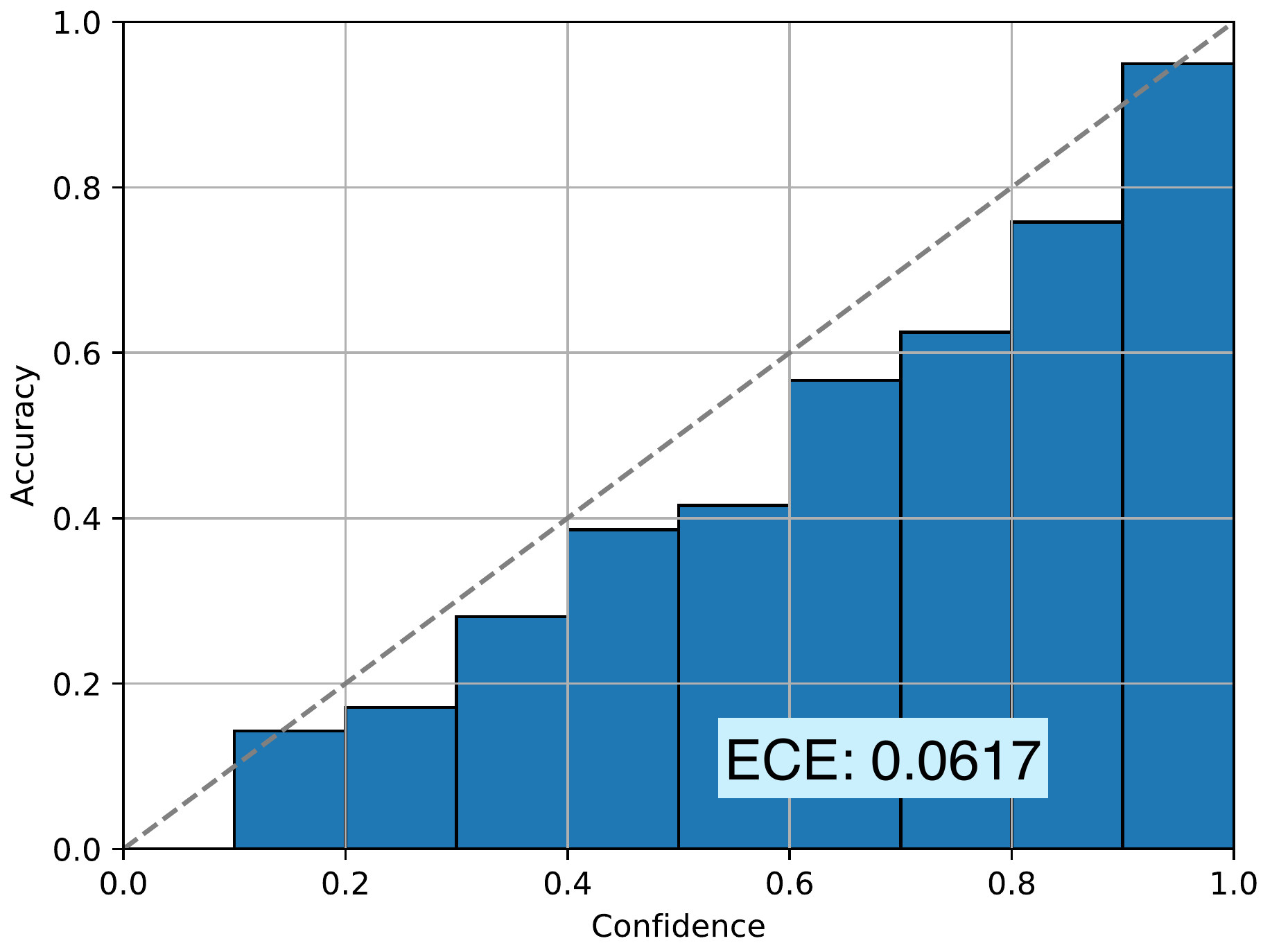}
\end{subfigure}
\caption{\footnotesize Reliability diagrams for DBSN, NEK-FAC, and MC dropout on CIFAR-10 (top row) and CIFAR-100 (bottom row). The bars aligning more closely to the diagonal are preferred. Smaller ECE is better.}\vspace{-0.cm}
\label{fig:ece}
\end{figure*}

\vspace{-0.1cm}
\label{app:cali}
We plot the reliability diagrams of 3 typical methods, which represent BNN with structure uncertainty, BNN with weight uncertainty, and deterministic NN with MC dropout, respectively, in Figure~\ref{fig:ece}. Obviously, DBSN has better reliability diagrams than NEK-FAC and MC dropout, proving the effectiveness of the uncertainty on network structure.

\vspace{-0.1cm}
\subsection{The Predictive Uncertainty for Adversarial Examples and Out-of-distribution (OOD) Samples}
\vspace{-0.1cm}
\label{app:pred-unc}
We further compare DBSN with the baselines by inspecting their
predictive uncertainty for adversarial examples and OOD samples.

\begin{figure}[t]
\centering
\begin{subfigure}[b]{0.255\columnwidth}
\centering
    \includegraphics[width=\textwidth]{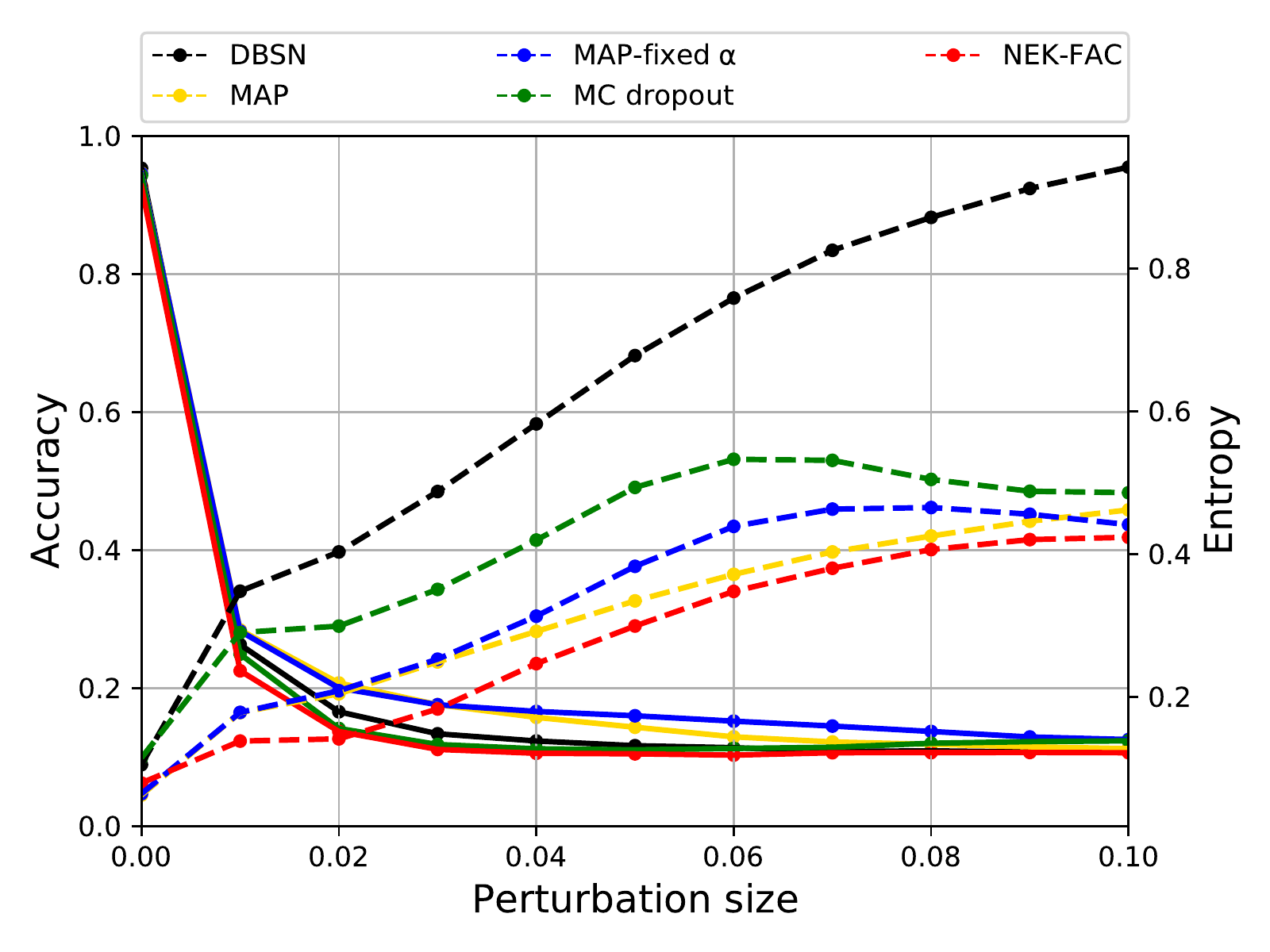}
    \vspace{-3.7ex}
    \caption{}
    
    \label{fig:1x}
\end{subfigure}
\begin{subfigure}[b]{0.255\columnwidth}
\centering
    \includegraphics[width=\textwidth]{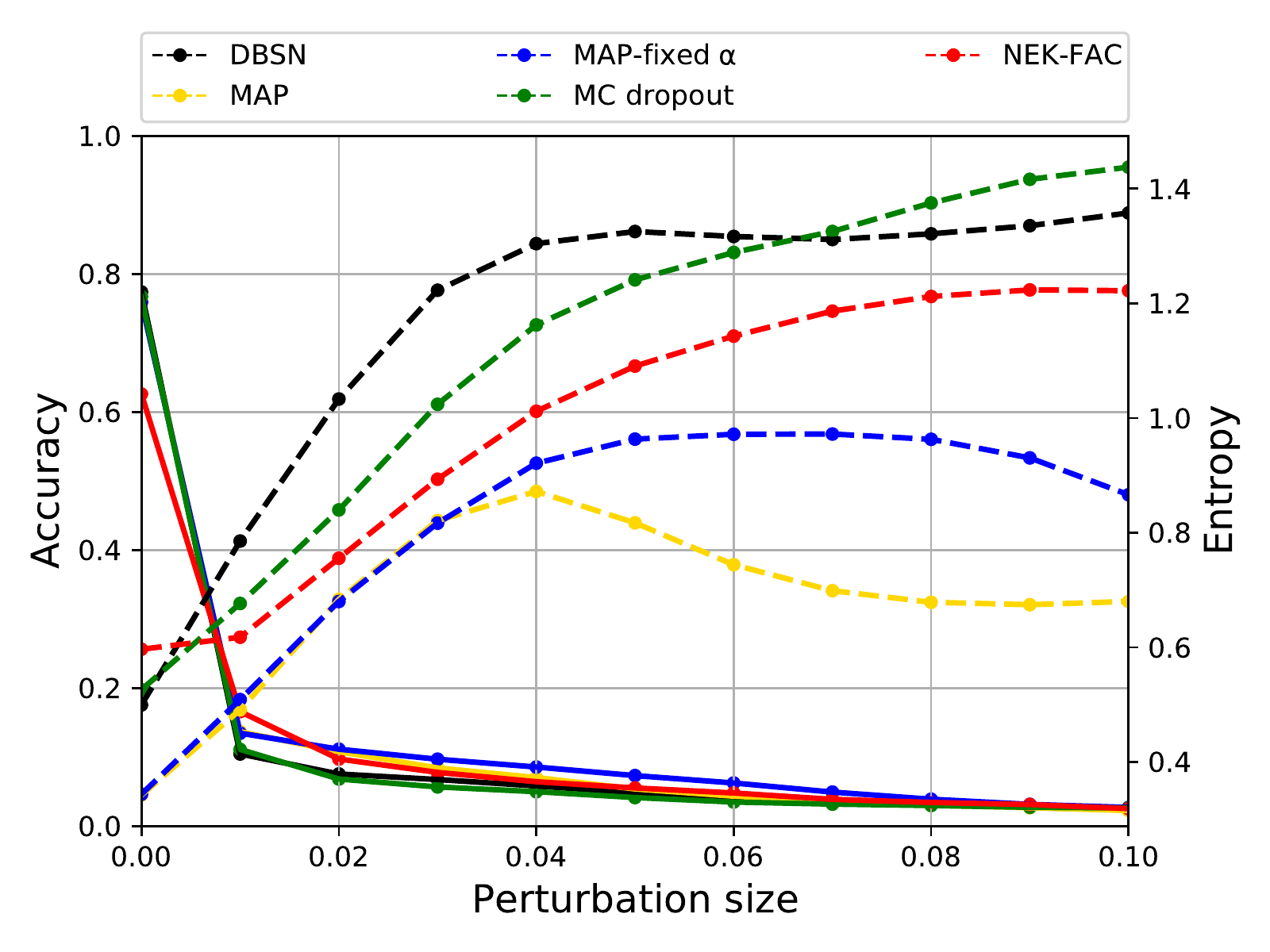}
    \vspace{-3.7ex}
    \caption{}
    \label{fig:1a}
\end{subfigure}
\begin{subfigure}[b]{0.225\columnwidth}
\centering
    \includegraphics[width=\textwidth]{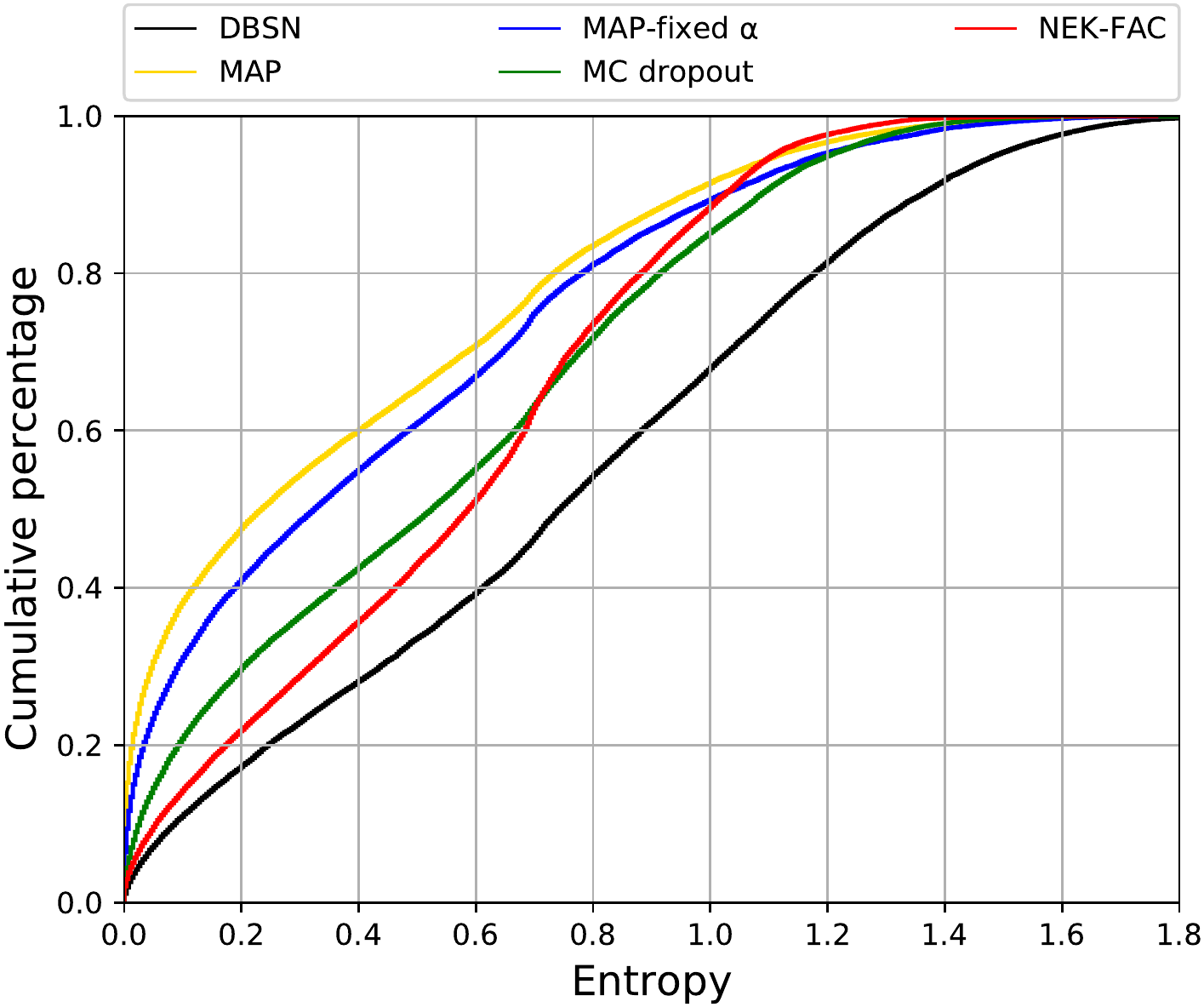}
    \vspace{-3.7ex}
    \caption{}
    \label{fig:1b}
\end{subfigure}
\begin{subfigure}[b]{0.225\columnwidth}
\centering
    \includegraphics[width=\textwidth]{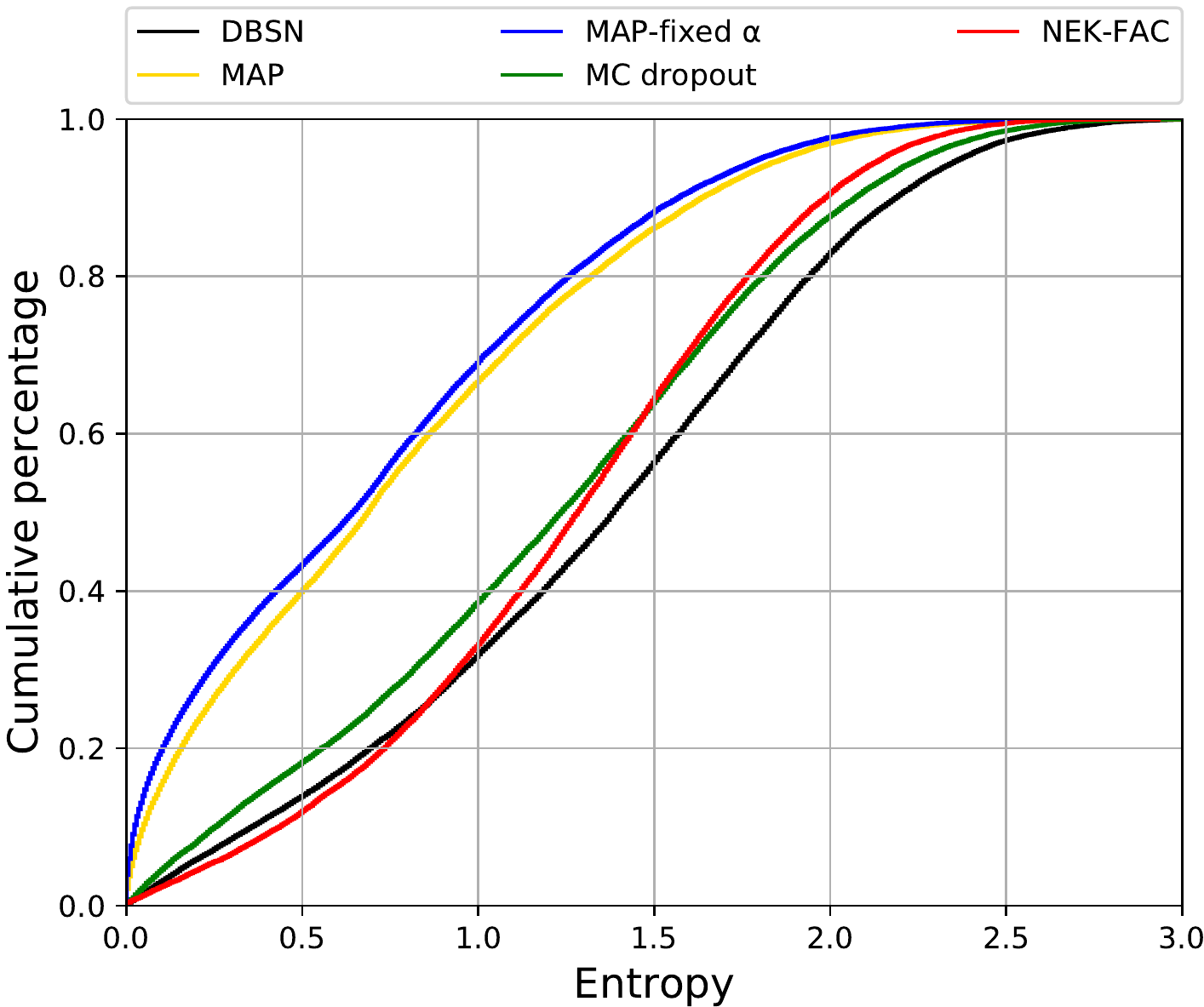}
    \vspace{-3.7ex}
    \caption{}
    \label{fig:1y}
\end{subfigure}
        \caption{\footnotesize (a)(b): accuracy (solid) and entropy (dashed) vary \emph{w.r.t.} the adversarial perturbation size on CIFAR-10 and CIFAR-100, respectively. (c)(d): empirical CDF for the predictive entropy of OOD samples on CIFAR-10 and CIFAR-100, respectively (curves closer to the bottom right corner are better).}
    \label{fig:pu-cifar10}
    \vspace{-0.cm}
\end{figure}

\textbf{Predictive uncertainty for adversarial examples.} Suggested by existing works~\citep{louizos2017multiplicative,pawlowski2017implicit}, we take the predictive entropy as a proxy of predictive uncertainty.
Concretely, we apply the frequently adopted  fast gradient sign method (FGSM)~\citep{goodfellow2014explaining} to attack the posterior predictive distribution of the trained models on CIFAR-10 and CIFAR-100. 
Then we calculate the predictive entropy for the crafted adversarial examples and depict the average entropy \emph{w.r.t.} the adversarial perturbation size in Figure~\ref{fig:pu-cifar10}(a)(b). 
We do not include {BBB} and {FBN} into comparison owing to their compromising performance and calibration.
As expected, the entropy of DBSN grows rapidly as perturbation size increases, implying DBSN becomes pretty uncertain when encountering adversarial perturbations. 
By contrast, the change in the entropy of MC dropout and NEK-FAC is relatively moderate, showing less sensitivity to the adversarial examples.

\begin{wrapfigure}{r}{0.5\linewidth}
\centering
\vspace{-.0ex}
\includegraphics[width=0.95\linewidth]{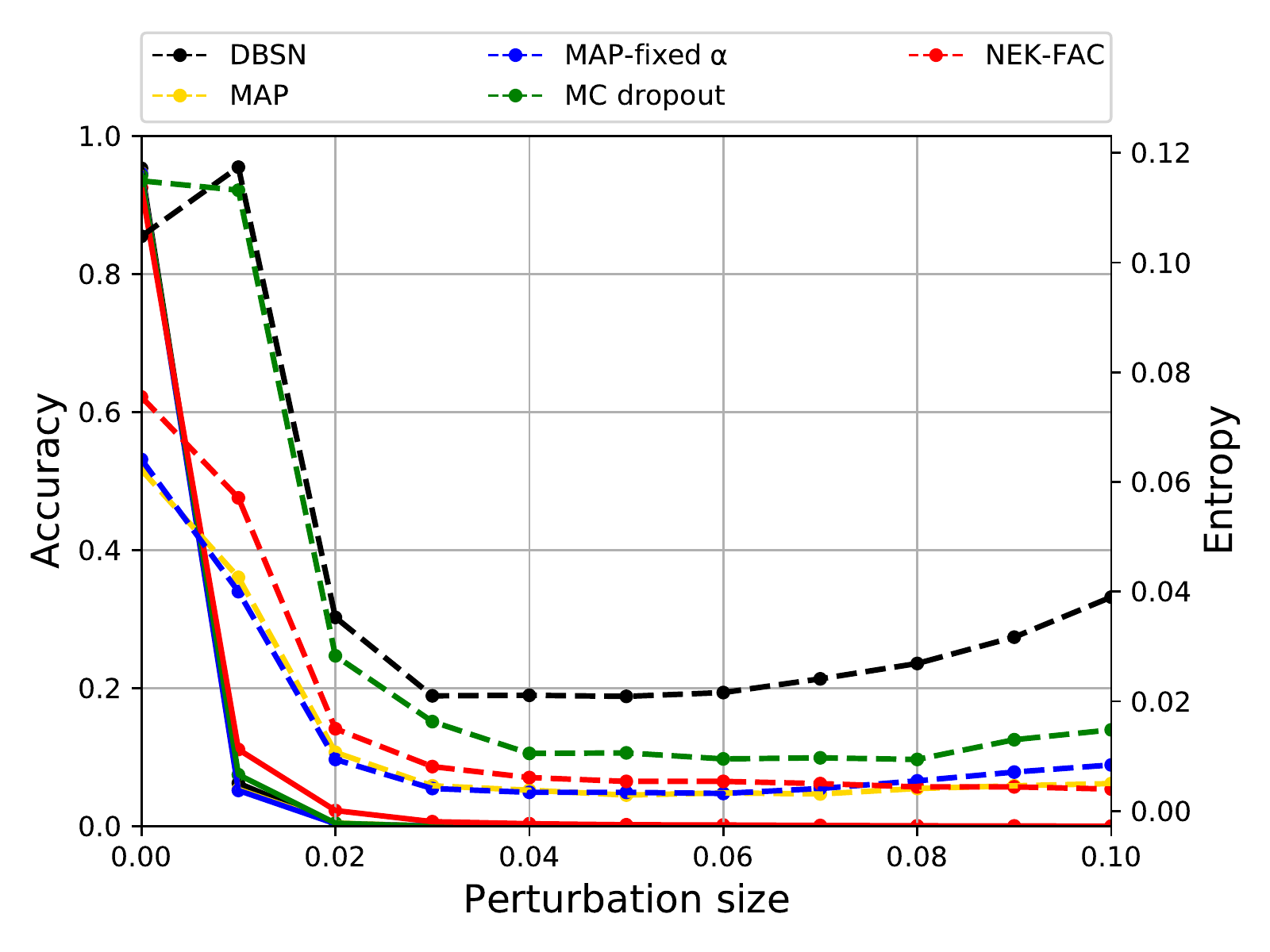}
\vspace{-.9ex}
\caption{\footnotesize Accuracy (solid) vs entropy (dashed) as a function of the adversarial perturbation size on CIFAR-10. Attacked by BIM.}
\vspace{-5ex}
\label{fig:adv-bim}
\end{wrapfigure}
We further attack with the more powerful  Basic Iterative Method (BIM)~\citep{kurakin2016adversarial}.
We set the number of iteration to be 3 and set the perturbation size in every step to be 1/3 of the whole perturbation size. The experiments mainly focus on the models trained on CIFAR-10. 
Figure~\ref{fig:adv-bim} shows the results. DBSN has increasing entropy when the perturbation size changes from 0 to 0.01, but all the other approaches are attacked successfully with entropy dropping. However, we have to agree that {BIM is powerful enough to break all the methods, including DBSN}. So we advise adjusting DBSN accordingly (\emph{e.g.}, employing adversarial training, using more robust loss) if we want to use DBSN to defend the adversarial attacks.

\textbf{Predictive uncertainty for out-of-distribution samples.} One further step, we look into the predictive entropy of OOD samples, to adequately evaluate the quality of uncertainty estimates. 
We use the trained models on CIFAR-10 and CIFAR-100 to predict the test data of SVHN. 
We calculate their predictive entropy and draw the empirical CDF in Figure~\ref{fig:pu-cifar10}(c)(d), following \citep{louizos2017multiplicative}. 
The curve close to the bottom right corner is expected as it means most OOD samples have relatively large entropy (\emph{i.e.}, low prediction confidence). 
Obviously, DBSN is less susceptible to OOD samples than the baselines.

\begin{figure*}[t]
\vspace{-0.cm}
\centering
\begin{subfigure}{0.32\textwidth}
  \centering
  \includegraphics[width=\linewidth]{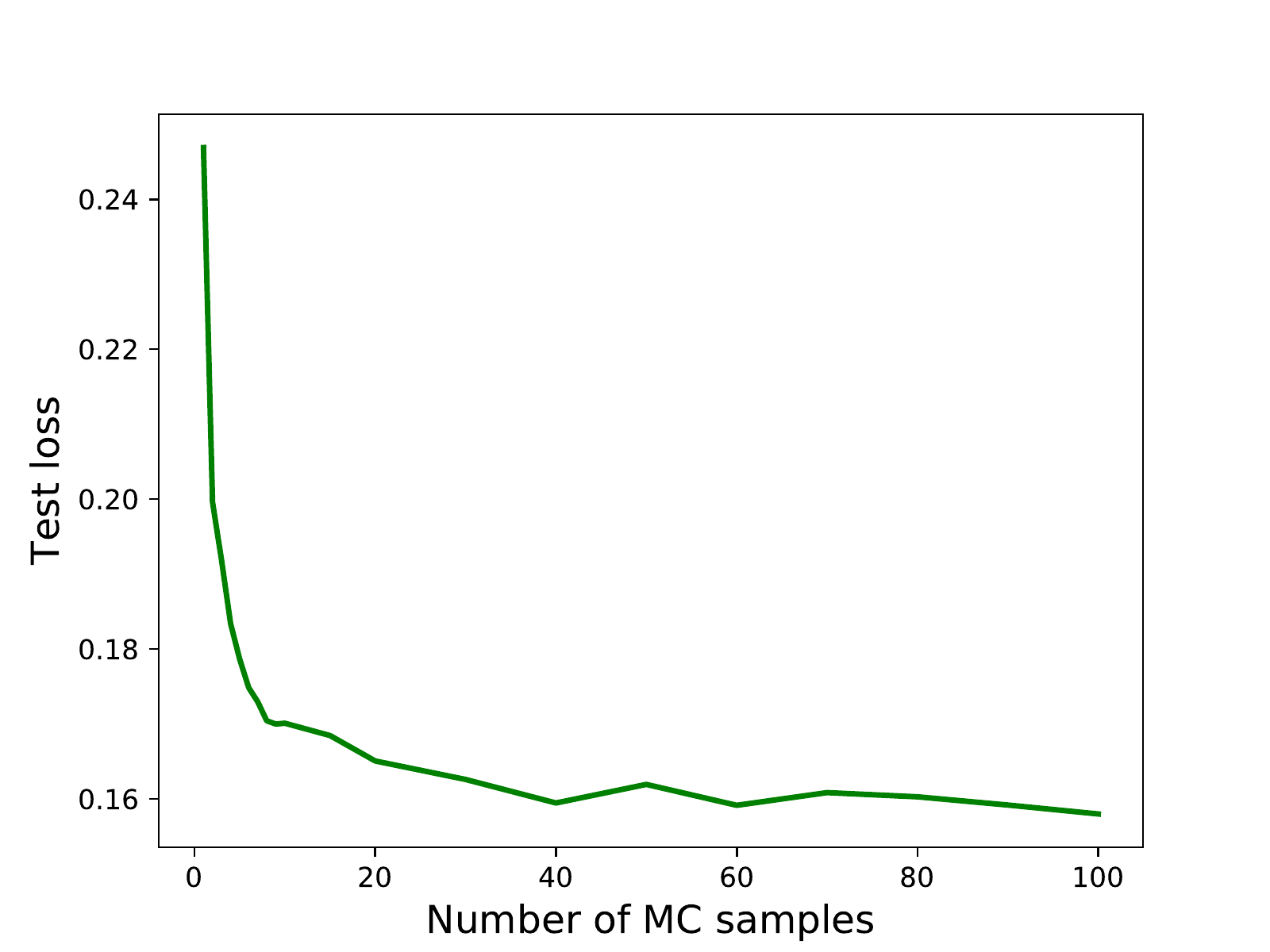}
\end{subfigure}
\begin{subfigure}{0.32\textwidth}
  \centering
  \includegraphics[width=\linewidth]{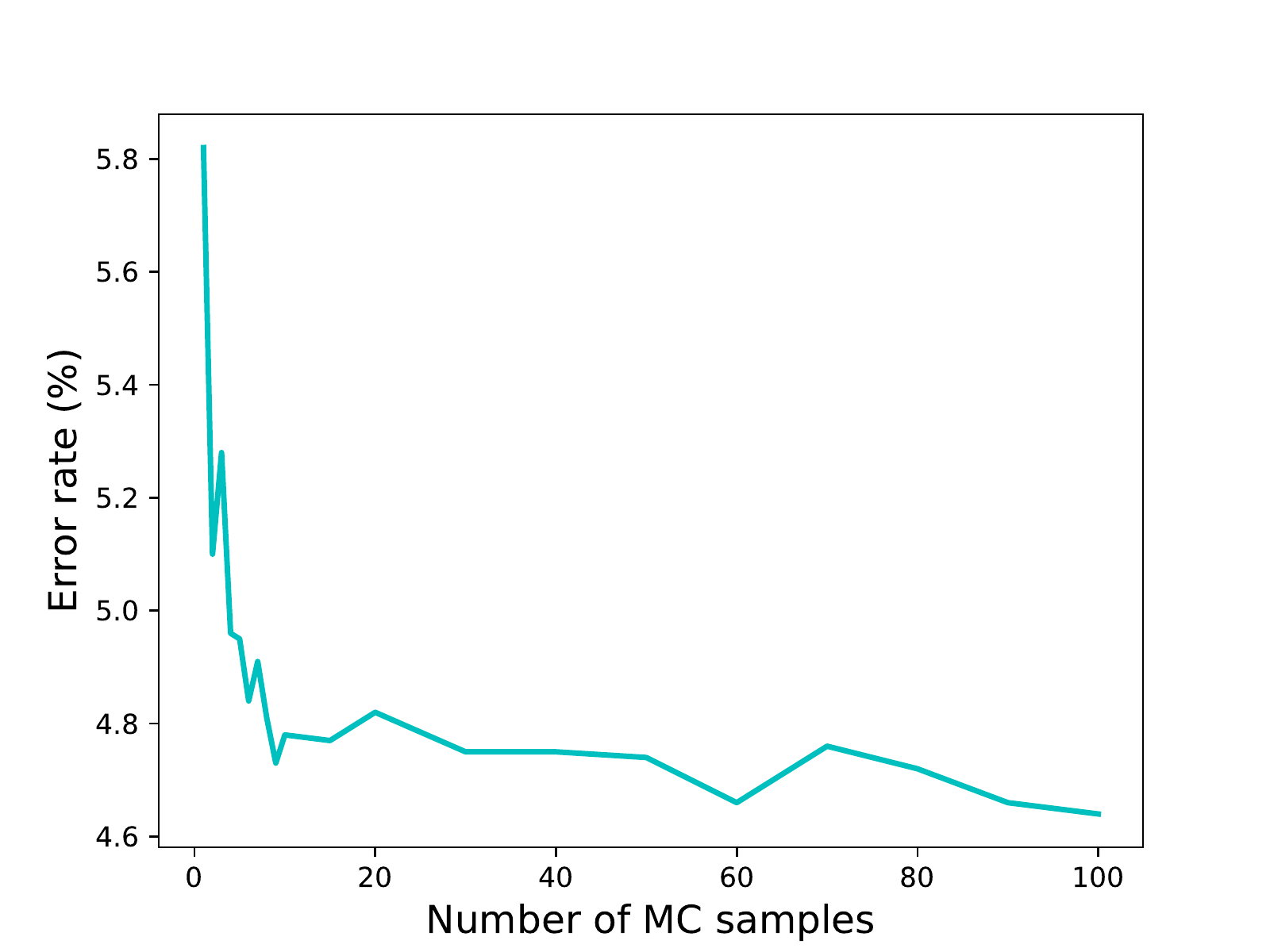}
\end{subfigure}
\begin{subfigure}{0.32\textwidth}
  \centering
  \includegraphics[width=\linewidth]{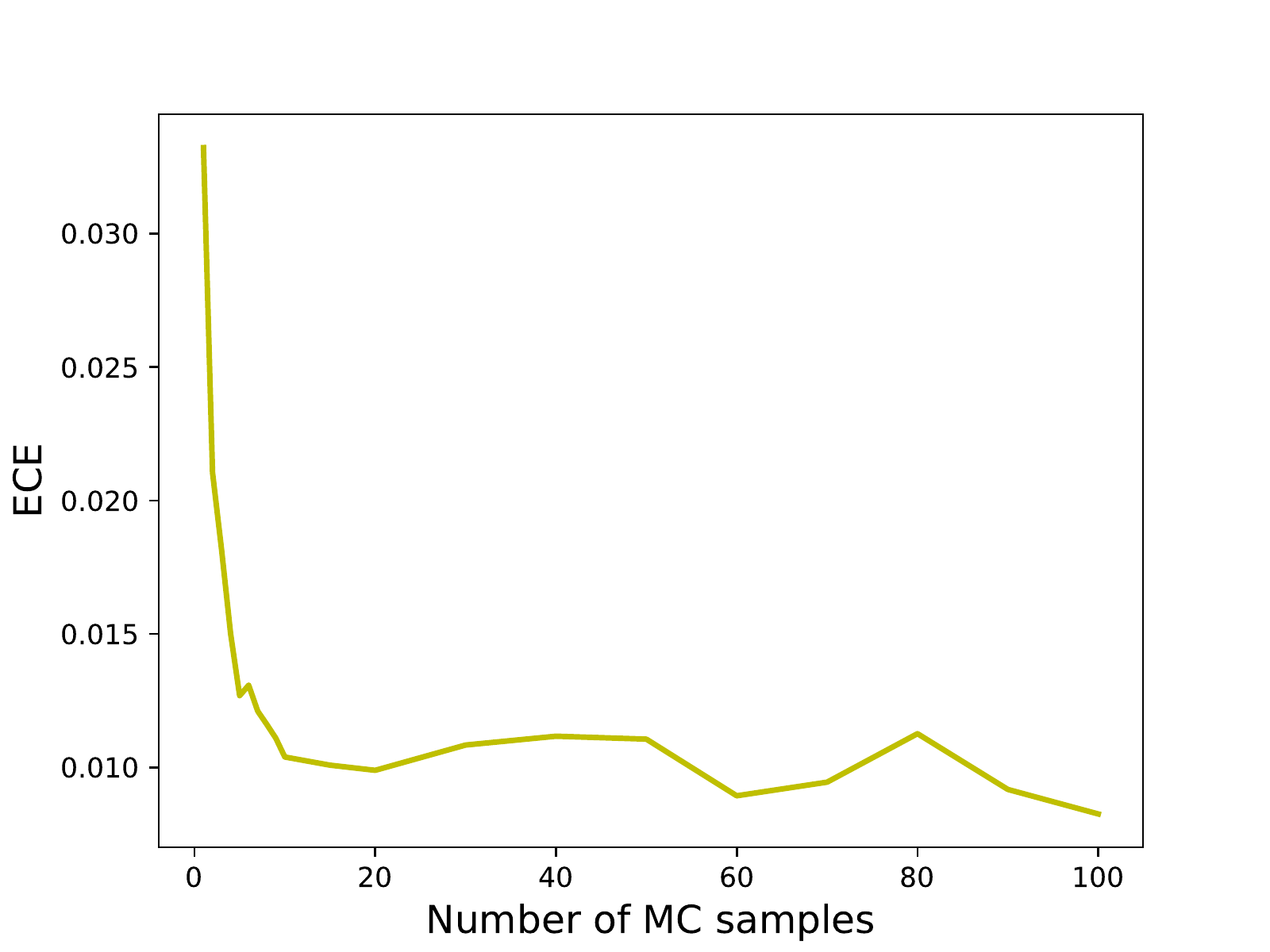}
\end{subfigure}
\vspace{-0.1cm}
\caption{\red{\footnotesize Test loss (left), test error rate (middle), and test ECE (right) of DBSN vary w.r.t. the number of MC samples used in \emph{Bayes ensemble}. (CIFAR-10)}}\vspace{-0.cm}
\label{fig:eff_mc_10}
\end{figure*}

\begin{figure*}[t]
\vspace{-0.cm}
\centering
\begin{subfigure}{0.32\textwidth}
  \centering
  \includegraphics[width=\linewidth]{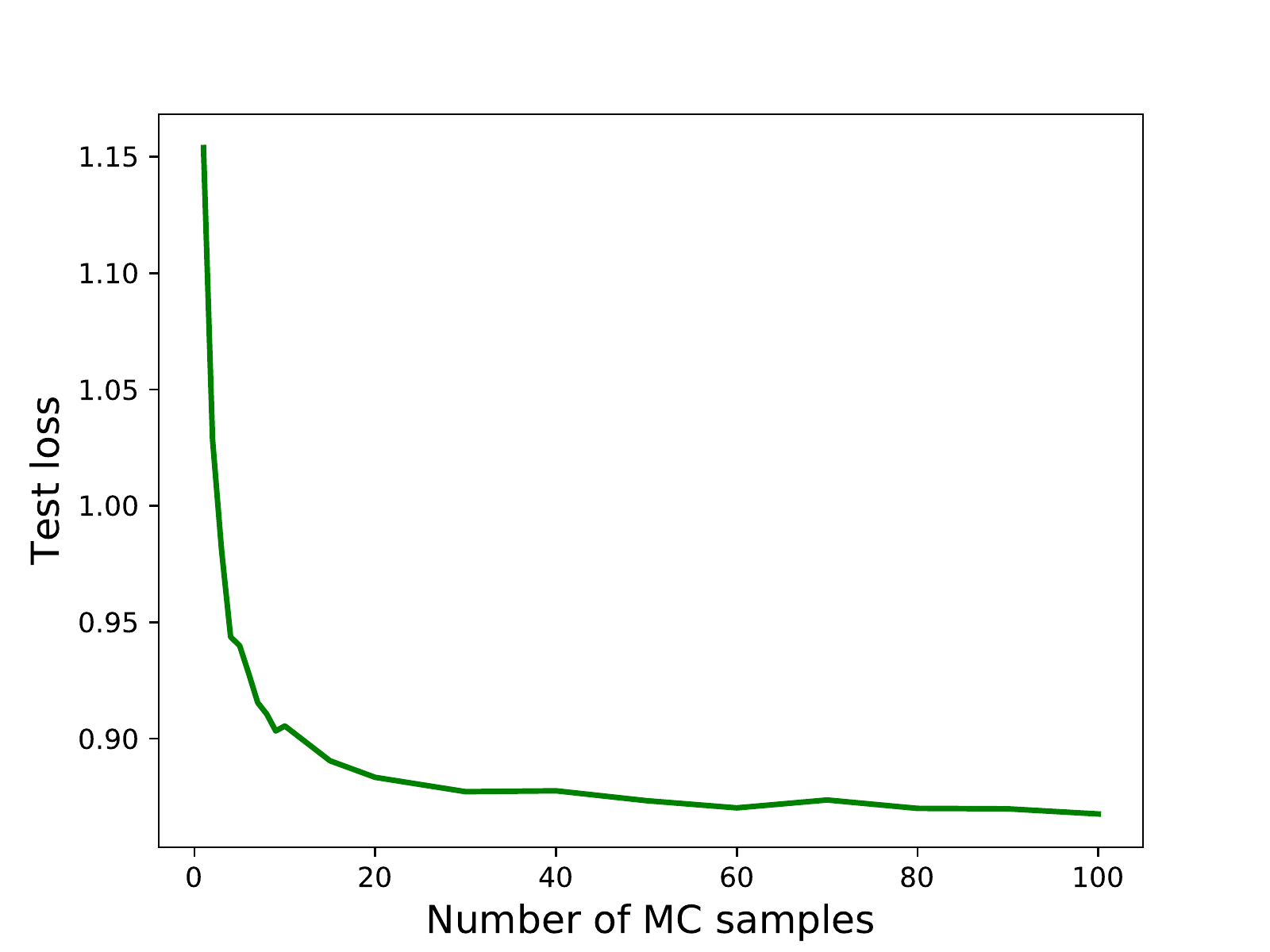}
\end{subfigure}
\begin{subfigure}{0.32\textwidth}
  \centering
  \includegraphics[width=\linewidth]{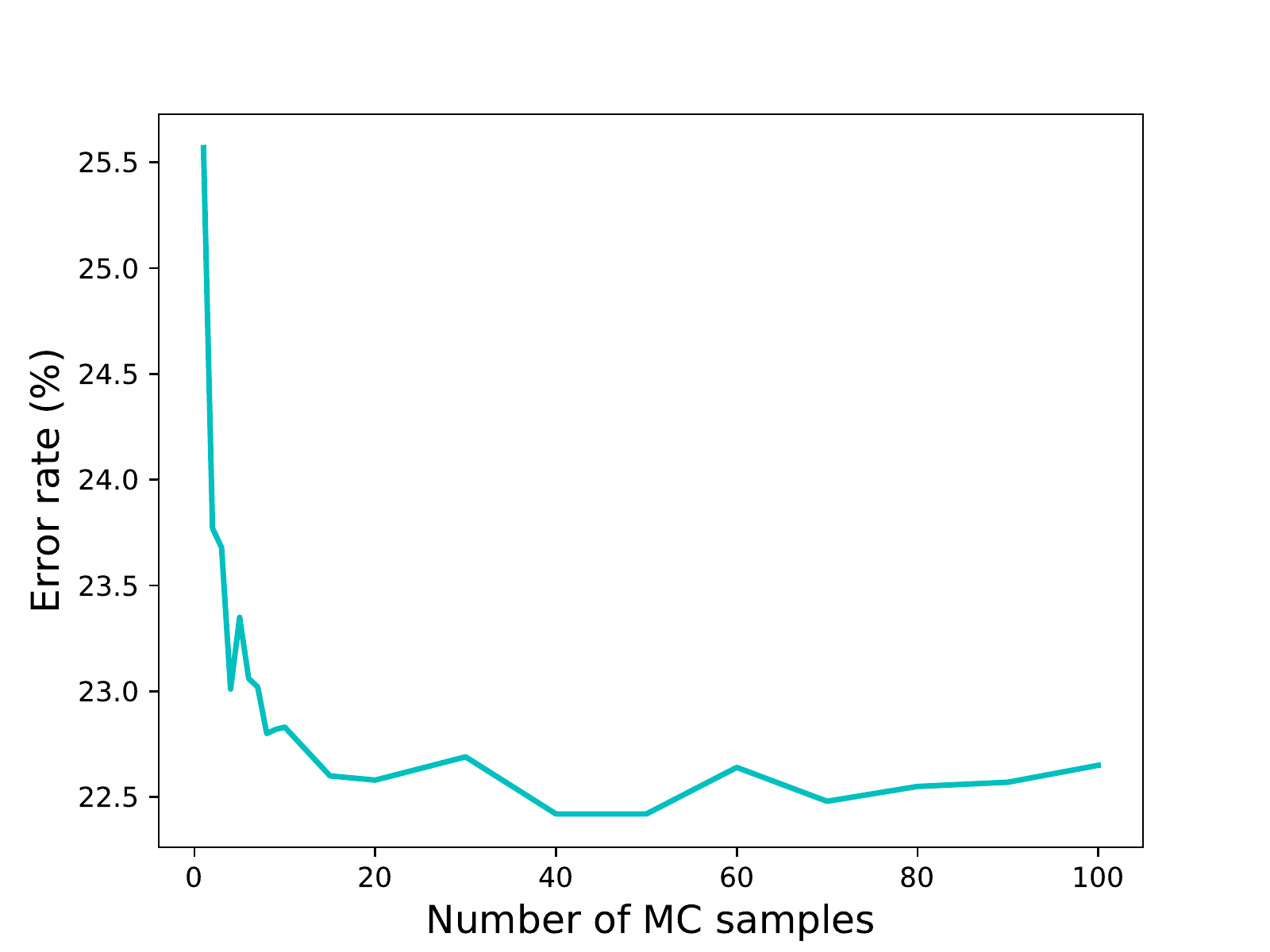}
\end{subfigure}
\begin{subfigure}{0.32\textwidth}
  \centering
  \includegraphics[width=\linewidth]{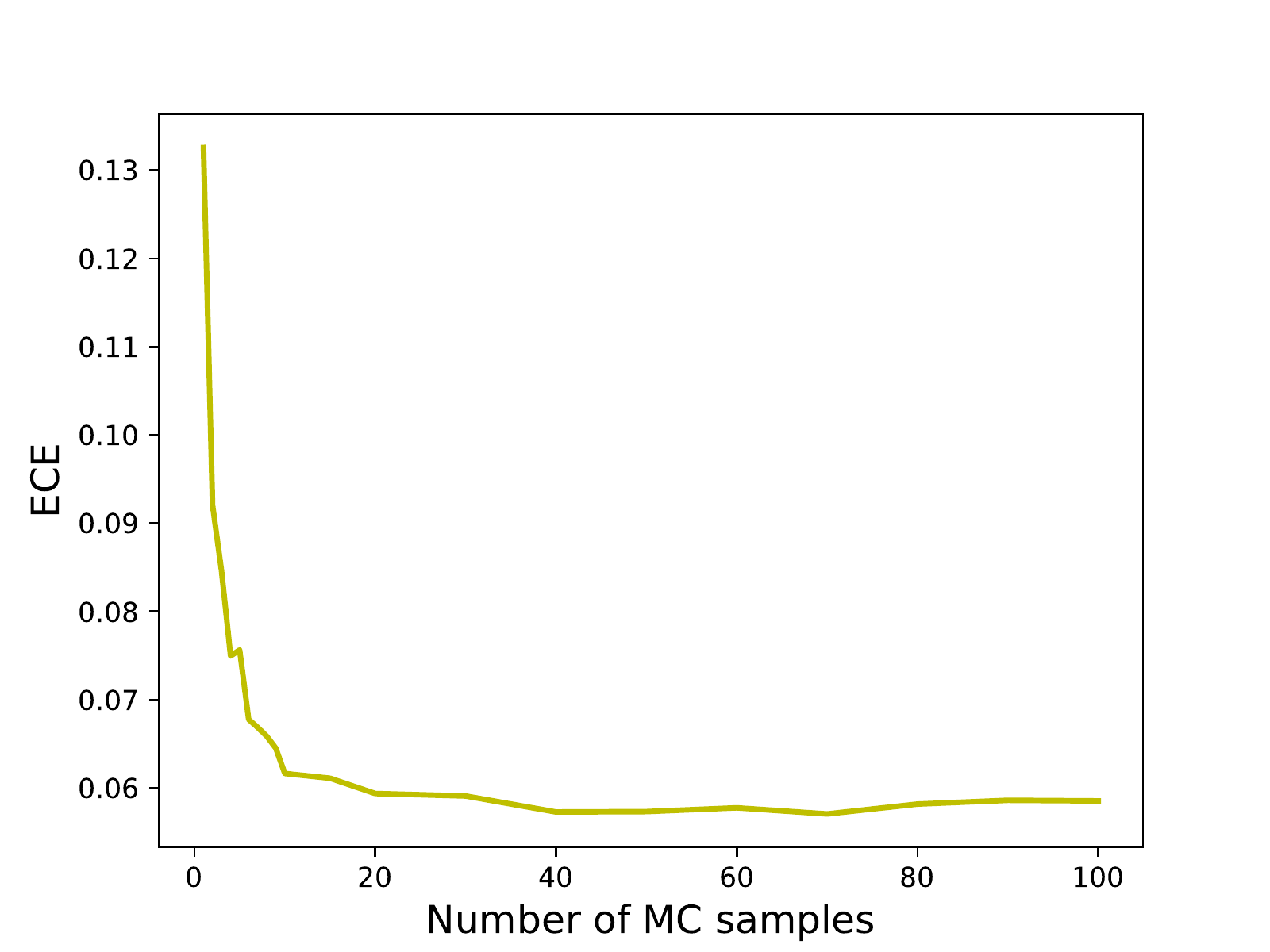}
\end{subfigure}
\vspace{-0.1cm}
\caption{\red{\footnotesize Test loss (left), test error rate (middle), and test ECE (right) of DBSN vary w.r.t. the number of MC samples used in \emph{Bayes ensemble}. (CIFAR-100)}}\vspace{-0.cm}
\label{fig:eff_mc_100}
\end{figure*}

\subsection{\red{The Effects of the Number of MC Samples for Bayes Ensemble}}
\label{app:eff_mc}
We draw the change of test loss, test error rate and test ECE with respect to the number of MC samples used by DBSN for \emph{Bayes ensemble} in Figure~\ref{fig:eff_mc_10} (CIFAR-10) and Figure~\ref{fig:eff_mc_100} (CIFAR-100). It is clear that assembling the predictions from models with various sampled network structures enhances the final predictive performance and calibration significantly. 
As shown in the plots, we would better utilize 20+ MC samples to predict the unseen data, for adequately exploiting the learned structure distribution. Indeed, we use 100 MC samples in all the experiments, except for the adversarial attack experiments where we use 30 MC samples for attacking and evaluation.

\end{document}